%% file: iclr2026_conference.tex
\documentclass{article} 
\usepackage{iclr2026_conference,times}

\input{math_commands.tex}

\usepackage{hyperref}
\usepackage{url}

\usepackage[T1]{fontenc}

\usepackage{booktabs}
\usepackage{makecell}
\usepackage{graphicx}
\usepackage{longtable}

\title{Toward Cross-Lingual Quality Classifiers for Multilingual Pretraining Data Selection}

\author{Yassine Turki, Vinko Sabolčec, Bettina Messmer, \& Martin Jaggi\\ 
Machine Learning Optimization Lab\\
Ecole Polytechnique Fédérale de Lausanne (EPFL)\\
Lausanne, Switzerland \\
\texttt{\{yassine.turki,vinko.sabolcec,bettina.messmer,martin.jaggi\}@epfl.ch} \\
}

\iclrfinalcopy 
\begin{document}

\maketitle

\begin{abstract}
As Large Language Models (LLMs) scale, data curation has shifted from maximizing volume to optimizing the signal-to-noise ratio by performing quality filtering. However, for many languages, native high-quality data is insufficient to train robust quality classifiers. This work investigates the idea that quality markers in embedding space may show cross-lingual consistency, which would allow high-resource languages to subsidize the filtering of low-resource ones.
We evaluate various filtering strategies, including cross-lingual transfer, third quartile sampling (Q3), and retention rate tuning. Our results demonstrate that massive multilingual pooling frequently outperforms monolingual baselines in both rank stability and aggregate accuracy for a 1B model trained on 103B tokens, delivering gains for high resource languages (1.2\% increase in aggregate normalized accuracy for French) and matching or exceeding monolingual baselines for low-resource languages. However, we find that scale alone does not guarantee stability. Furthermore, for high-resource languages like French, we show that refining the decision boundary through third quartile sampling (Q3) or tuning the retention rate is necessary to fully leverage the multilingual signal.
\end{abstract}

\section{Introduction}
As machine learning architectures and optimization strategies have matured, the focus of the community has shifted toward the foundational element of model performance: data quality. Previous work has demonstrated that ``more'' is not synonymous with ``better'' \citep{c4, refinedweb}. Building on this insight, recent efforts have shown that aggressively curating high-quality subsets from web-scale corpora can match or exceed the performance of models trained on far larger datasets \citep{dclm, finewebedu, fineweb2HQ}. 

Historically, data curation relied on heuristic-based filtering~\citep{culturax, rl1, rl2, refinedweb}. However, the emergence of model-based classifiers has enabled a more nuanced selection of semantically dense content. FineWeb-Edu~\citep{finewebedu} reached the performance of LLMs trained on $350B$ tokens of raw English data using only 10\% of tokens through LLM-based filtering. Building upon this, FineWeb2-HQ~\citep{fineweb2HQ} extended model-based filtering to multilingual data, training classifiers for 20 languages and demonstrating that 15\% of tokens could match the performance of models trained on the full FineWeb2~\citep{fineweb2} dataset.

While model-based filtering has proven effective for high-resource languages, the multilingual domain faces a critical challenge: many languages lack sufficient native high-quality data to train effective standalone classifiers. This work investigates whether quality classifiers can generalize across languages by exploiting shared semantic structures in multilingual embedding spaces, which would enable high-resource languages to effectively support filtering for low-resource ones.


We hypothesize that quality is a measurable density of information and logical coherence rather than subjective preference.
We assume high-quality text may be distinguished in the embedding space by \textbf{grammatical coherence}, where formal structures create distinct activation patterns; \textbf{lexical density}, characterized by specialized vocabulary rather than boilerplate; and \textbf{information density}, derived from high logical and factual coherence. If this hypothesis holds, a classifier $\mathcal{C}$ trained to recognize quality patterns in French should transfer to typologically distant languages like Chinese, either because they share an underlying quality manifold, or because structural and formatting regularities in the positive anchor datasets (such as Wikipedia markup or instruction-tuning templates) are themselves cross-lingual proxies for quality. 
Our primary contributions are as follows:
\begin{itemize}
\item \textbf{Multilingual Synergy:} We demonstrate that massive multilingual pooling frequently outperforms monolingual baselines in both rank stability and aggregate accuracy, delivering gains for high-resource languages and beating state-of-the-art baselines for low-resource ones.
\item \textbf{Empirical Validation of Cross-Family Transfer:} We demonstrate that classifiers trained on one language family can effectively curate high-quality tokens in typologically distant languages.
    \item \textbf{Q3 Sampling Strategy:} We introduce a third-quartile (Q3) sampling strategy that trains against fluent but low-utility text. We find this refinement sharpens decision boundaries and improves performance in high-resource languages such as French and Spanish.
\end{itemize}

\section{Related Work}

\paragraph{Evolution of Web Data Curation.} Early large-scale datasets relied primarily on heuristic-based filtering of Common Crawl. \citet{ccnet} introduced CCNet, which utilized FastText~\citep{joulin2016bagtricksefficienttext} for language identification and perplexity-based scoring. This approach was refined by \citet{c4} with C4 and \citet{refinedweb} with RefinedWeb, the latter demonstrating that aggressive deduplication and string-matching heuristics could allow web data to match curated dataset performance. However, \citet{finewebedu} showed that simple heuristics fail to capture the nuanced educational value required for complex reasoning tasks.

\paragraph{Cross-lingual Representation Learning}
Modern NLP has moved beyond language-specific models toward unified multilingual encoders. Models such as XLM-RoBERTa~\citep{conneau2020unsupervisedcrosslingualrepresentationlearning} utilize a Transformer architecture trained on a masked language modeling objective across 100+ languages simultaneously.
The core power of these models lies in their ability to align semantic clusters across languages within a shared embedding space. During pre-training, the model learns that semantically equivalent words occupy similar topological positions regardless of surface form. For example, the English word ``Science'' and the German word ``Wissenschaft'' are positioned near each other in the 768-dimensional embedding space, as both relate similarly to concepts like ``logic'' and ``fact.'' This alignment enables classification based on geometric position in latent space rather than surface-level syntax.

\paragraph{Model-Based Quality Filtering.} A paradigm shift occurred with model-based classifiers. FineWeb-Edu \citep{finewebedu} used Llama-3~\citep{grattafiori2024llama3herdmodels} as a judge to score educational quality and create a knowledge-rich dataset. To scale this approach, \citet{dclm} and \citet{fineweb2HQ} used lightweight classifiers based on FastText~\citep{joulin2016bagtricksefficienttext} or MLPs trained on embeddings of an XLM-RoBERTa model. Our work builds directly upon the FineWeb2-HQ pipeline by \citet{fineweb2HQ}.

\paragraph{Multilingual Scaling.} While English-centric curation is well-established, multilingual curation presents unique challenges. \citet{culturax}, \citet{madlad400} and \citet{fineweb2} expanded web-scale cleaning to hundreds of languages using perplexity and basic heuristics. FineWeb2-HQ~\citep{fineweb2HQ} advanced this by applying model-based filtering to 20+ languages. Despite these advances, two areas remain underexplored. First, while multilingual encoders are widely used, the degree to which one classifier can generalize across language families (e.g., from Nordic to Romance) has not been systematically characterized. Second, prior work primarily uses random negative sampling; the potential of smarter sampling techniques to refine decision boundaries remains largely uninvestigated. Our work addresses
both gaps.

\section{Methods}

\subsection{Classifier Datasets}
\label{sec:datasets}
To train a robust multilingual quality classifier, we curate a diverse set of high-quality ``positive'' samples and contrast them against a baseline of general web data. Our strategy extends the FineWeb2-HQ framework by expanding their high-quality anchors (\textbf{MKC+}) with additional instruction-based and synthetic sources to form the \textbf{MKC-e} (Extended) dataset.

\paragraph{Positive Anchors (MKC+).}
We incorporate the original FineWeb2-HQ anchors, which prioritize structured, knowledge-dense content. These datasets are Multilingual MMLU~\citep{mmmlu}, Aya Dataset and Collection~\citep{ayadataset},  OpenAssistant-2~\citep{openassist} and Include-Base-44~\citep{include}.

\paragraph{Extended Positives (MKC-e).}
To generalize the classifier's ability to recognize natural queries and encyclopedic prose, we expand the MKC+ pool with:

\begin{itemize}
    \item \textbf{Tagengo~\citep{tagengo}:} A multilingual chat dataset containing approximately 75,000 conversations in 74 languages between humans and GPT4~\citep{openai2024gpt4technicalreport}.
    \item \textbf{MURI-IT (Wikipedia Subset)~\citep{muri}:} A dataset containing instruction-output pairs across 200 languages. We specifically extracted the \textbf{Wikipedia subset} to ensure samples reflect factual, encyclopedic prose and are knowledge-rich. Furthermore, MURI-IT contains samples combining different languages. As our goal is to ablate classifier behaviour for different languages, we decided not to include multiple languages in a given sample. 
    \item \textbf{EuroBlocks-SFT-Synthetic~\citep{euroblocks}:} Multilingual synthetic data for Supervised Fine-Tuning used to train the EuroLLM 9B Instruct model. It spans 35 languages. 
    \item \textbf{WikiQA~\citep{apertus2025apertusdemocratizingopencompliant}}: A dataset linking real-world user queries to factual Wikipedia answer sentences. With 65 languages, it provides a stronger focus on low-resource languages. 
\end{itemize}

\paragraph{Negative Anchors (FineWeb2).}
We use the raw FineWeb2~\citep{fineweb2} corpus as our source of negative samples. We assume a random sample from this web-scale crawl primarily contains boilerplate, informal prose, or noise.

\subsection{Sampling and Data Preparation}
\paragraph{Balancing and Preprocessing.}
To ensure consistency, all samples are processed into 768-dimensional embeddings using the XLM-RoBERTa encoder~\citep{conneau2020unsupervisedcrosslingualrepresentationlearning} used in the original FineWeb2-HQ release. Consistent with prior work, we perform minimal preprocessing (concatenation of prompt/response) and remove samples with \texttt{<unk>} tokens.

For training, we sample $100,000$ positive documents per language, an increase from the $80,000$ used in~\citep{fineweb2HQ} to improve representation. To counter class imbalance in low-resource languages, we upsample positives by a maximum factor of $3$ (analogous to \citep{unimax}), ensuring high-resource languages do not dominate the gradient without overfitting to duplicated samples. We sample an equal number of negative documents to maintain a balanced class distribution.
For our classifier, we use a simple MLP, with a single hidden layer (256 dim, ReLU, 20\% dropout) and sigmoid output, trained to predict positive and negative classes based on XLM-RoBERTa embeddings in the same way as FineWeb2-HQ.

\paragraph{Negative Sampling Strategies.}
We employ two distinct strategies for selecting negative samples from FineWeb2:
\begin{enumerate}
    \item \textbf{Random Sampling:} The standard approach, selecting documents uniformly at random to distinguish quality content from general web noise.
    \item \textbf{Q3 (Hard Negatives):} To sharpen the decision boundary, we sample documents that score in the 50th--75th percentile (the third quartile) of a preliminary classifier. These "Q3 negatives" typically represent fluent but low-utility text (e.g., repetitive procedural content), forcing the model to learn subtler distinctions beyond surface-level fluency.
\end{enumerate}

\section{Evaluation}
\paragraph{Qualitative Analysis.}


To understand how each classifier filters samples, we first verify that score distributions concentrate near zero with a flatter tail at higher scores. We then inspect the top and bottom 25 samples to confirm that highly-ranked samples are knowledge-rich and well-structured, while low-ranked ones are poorly written or uninformative. Using a monolingual FineWeb2-HQ classifier as baseline, we compute rank correlations (Spearman and Kendall) and analyze the top 20 samples with the largest rank increases and decreases across classifiers. This reveals the filtering patterns each classifier prioritizes (e.g., grammar, content quality, symbols). Selected ranking changes are shown in Appendix~\ref{appendix:samples}.

\paragraph{Downstream Task Evaluation.}
To better understand how different filtering methods would influence the performance of an LLM, we conduct experiments by training a small 1B parameter Apertus~\citep{apertus2025apertusdemocratizingopencompliant} architecture model on the filtered FineWeb2 by a given classifier. The model is trained on 103B tokens with a sequence length of $4096$, and sees each token at most twice, except for Arabic, where for retention rates of 10\% and 20\%, we have replicated the filtered data 10x and 5x respectively, to account for a lower number of tokens. These tokens are directly from the filtered samples of the classifier, followed by a rehydration step as described in~\citep{fineweb2}. Technical details for the LLM training can be found in Appendix~\ref{appendix:meg_config}.

To evaluate the models, we use the Language Model Evaluation Harness library~\citep{eval-harness}. Our main criterion is normalized accuracy, as recommended by~\citet{kydlicek2024finetasksmultilingualtasks}. The tasks we use span multiple capabilities, including knowledge retrieval, reasoning and natural language understanding. Full benchmark list can be found in Appendix~\ref{appendix:benchmarks}.
The average rank across these benchmarks serves as our final measure of a filtering strategy's robustness. We also report the mean normalized accuracy, as the average rank can be volatile for very small differences in performance.

\section{Experiments and Results}

We conduct ablations across four languages representing diverse linguistic profiles: French (Romance, high-resource), Spanish (Romance, high-resource), Arabic (Semitic, morphologically complex), and Chinese (Sino-Tibetan, logographic). Our experimental design addresses three core questions:

\begin{enumerate}
    \item \textbf{Multilingual Synergy:} Does pooling data from multiple languages improve performance compared to monolingual baselines?
    \item \textbf{Cross-lingual Transfer:} Can classifiers trained on typologically distant languages identify quality in a target language?
    \item \textbf{Decision Boundary Refinement:} Can the Q3 strategy improve classifier precision in high-resource settings?
\end{enumerate}

\paragraph{Baseline Configurations.} We establish two primary baselines: \textbf{No filtering}, representing random sampling from FineWeb2 (lower bound), and \textbf{HQ}, a monolingual classifier trained following the FineWeb2-HQ pipeline~\citep{fineweb2HQ} with MKC+ anchors.

\paragraph{Multilingual Synergy.}
\label{sec:CLT}
We first investigate whether data quality is a general feature shared across samples from different languages, or if it is language-specific. Our hypothesis is that there exist shared regularities in XLM-RoBERTa's embedding space that correlate with quality. If we train a model to detect these regularities across multiple languages, then we would obtain a general classifier that could recognize high-quality samples in unseen languages, and even boost the data quality for languages it was trained on. 
In order to test this hypothesis, we train a general multilingual classifier (denoted by \textbf{ML}). We use the \textbf{MKC-e} pool to account for the lack of samples for some knowledge, and also to balance the dominance of the Aya Collection dataset in terms of size and representation. The detailed counts for each language can be found in the appendix~\ref{sec:datacounts}.  

Tables~\ref{benchmark:chinese_ML} and~\ref{benchmark:spanish_ML} evaluate the effect of multilingual quality filtering on downstream performance for a 1B model in Chinese and Spanish. The multilingual classifier (ML) achieves the highest aggregate accuracy and lowest average rank in both languages, outperforming both No filtering and monolingual high-quality filtering (HQ).

Improvements are observed consistently across diverse reasoning and natural language understanding benchmarks, such as ARC~\cite{ARC}, GMMLU~\cite{singh2025globalmmluunderstandingaddressing}, XNLI~\citep{xnli}, and Include~\citep{include}. While HQ occasionally yields marginal gains on individual tasks, ML provides more stable and stronger overall performance.

These results are consistent with the hypothesis that data quality corresponds to a shared structure in XLM-RoBERTa's representation space. However, we cannot determine whether this reflects abstract semantic features or cross-lingual formatting regularities in the positive anchor datasets. By jointly modeling quality across languages, the classifier captures features that transfer effectively across linguistic boundaries, leading to improved downstream generalization.

\begin{table}[t]

\caption{Comparison of models trained without filtering, with monolingual high-quality (HQ) filtering, and with our multilingual (ML) classifier on Chinese. The ML strategy achieves the highest aggregate accuracy and best average rank, supporting the hypothesis that quality features are transferable.}
\label{benchmark:chinese_ML}

\begin{center}
\begin{tabular}{lccc}
\multicolumn{1}{c}{\bf Benchmark} &
\multicolumn{1}{c}{\bf No filtering} &
\multicolumn{1}{c}{\bf HQ} &
\multicolumn{1}{c}{\bf ML}\\
\midrule
Agieval Cn      & 0.3618 & \textbf{0.3644} & 0.3457 \\
ARC             & 0.2855 & 0.3145 & \textbf{0.3171} \\
Belebele\_c     & 0.3011 & 0.3200 & \textbf{0.3222} \\
Ceval-valid     & 0.2288 & 0.2489 & \textbf{0.2615} \\
Cmmlu\_c        & 0.3206 & 0.3471 & \textbf{0.3608} \\
GMMLU\_c        & 0.2800 & 0.3075 & \textbf{0.3200} \\
Include\_c      & 0.3468 & 0.3523 & \textbf{0.3541} \\
MMLU\_c         & 0.2772 & 0.2940 & \textbf{0.2987} \\
PAWS            & 0.5520 & 0.5535 & \textbf{0.5610} \\
Xcopa           & 0.5860 & 0.5920 & \textbf{0.6080} \\
XNLI            & 0.3546 & 0.4072 & \textbf{0.4189} \\
Xstorycloze     & 0.6559 & \textbf{0.6625} & 0.6625 \\
XWinograd       & 0.6806 & \textbf{0.6825} & 0.6766 \\
\hline
\textbf{Aggregate acc\_norm} & 0.4024 & 0.4190 & \textbf{0.4236} \\
\textbf{Average rank} & 2.85 & 1.77 & \textbf{1.31} \\
\hline
\end{tabular}
\end{center}

\end{table}

\begin{table}[t]

\caption{Comparison of models trained without filtering, with monolingual high-quality (HQ) filtering, and with our multilingual (ML) classifier on Spanish. The ML strategy achieves the highest aggregate accuracy and best average rank, supporting the hypothesis that quality features are transferable.}
\label{benchmark:spanish_ML}

\begin{center}
\begin{tabular}{lccc}
\multicolumn{1}{c}{\bf Benchmark} &
\multicolumn{1}{c}{\bf No filtering} &
\multicolumn{1}{c}{\bf HQ} &
\multicolumn{1}{c}{\bf ML}\\
\midrule
ARC-Challenge & 0.2991 & 0.3077 & \textbf{0.3248} \\
Belebele\_c   & 0.3422 & \textbf{0.3533} & 0.3456 \\
GMMLU\_c      & 0.3100 & 0.3150 & \textbf{0.3250} \\
HellaSwag     & 0.5006 & 0.5263 & \textbf{0.5310} \\
Include\_c    & 0.3491 & 0.3727 & \textbf{0.3891} \\
M\_MMLU\_c    & 0.2814 & 0.2985 & \textbf{0.3050} \\
XNLI          & 0.4618 & 0.4538 & \textbf{0.4783} \\
\hline
\textbf{Aggregate acc\_norm} & 0.3635 & 0.3753 & \textbf{0.3855} \\
\textbf{Average rank} & 2.86 & 2.00 & \textbf{1.14} \\
\hline
\end{tabular}
\end{center}

\end{table}

\paragraph{Cross-lingual Transfer.}
\label{subsec:nordic}
We have seen that multilingual transfer is possible since a classifier trained on multiple languages performs better than its monolingual counterpart. We note that our ML classifier was trained on these languages; therefore, it already has knowledge in that language. We hypothesize that this knowledge has been augmented by the data from the other languages. Can this ML classifier actually generalize to unseen languages? Would this classifier be able to perform as well (if not better) on a language like French if it has never seen a sample in that language?

\begin{table}[t]

\caption{Spearman and Kendall correlations between family-specific classifiers and the French HQ baseline. High correlation in distant families (e.g., Nordic) indicates a shared quality manifold, while the drop in "Romance (no French)" suggests potential syntactic interference.}
\label{tab:fra_retention_correlations}

\begin{center}
\begin{tabular}{lcc}
\multicolumn{1}{c}{\bf Experiment} &
\multicolumn{1}{c}{\bf Spearman} &
\multicolumn{1}{c}{\bf Kendall}\\
\midrule
Romance (spa, fra, por, ita, ron, cat) MKC+ & 0.8928 & 0.7173 \\
Nordic (swe, dan, nob, isl) MKC+            & 0.8820 & 0.6990 \\
Romance, no french (spa, por, ita, ron, cat) MKC-e & 0.7139 & 0.5228 \\
\hline
\end{tabular}
\end{center}

\end{table}

To answer these questions, we investigate the performance of a classifier on French. First, we train classifiers on different language families (i.e. Romance, Nordic, Germanic, etc.) and apply them on the French split of FineWeb2 to obtain scores. Then, we compute the Spearman and Kendall correlations between the scores of these classifiers and the HQ baseline. We would expect to have a very strong correlation score in the Romance family (French, Spanish, Portuguese, Italian, Romanian, Catalan) and a low correlation score in the other language families, which are linguistically distant from French. We display some of these correlation scores in Table~\ref{tab:fra_retention_correlations}, and provide the full table for other language families in the Appendix~\ref{appendix:rank_corr}. We can see that intra-family transfer is strong: Romance (MKC+) $\rho_s = 0.89$, indicating near-identical document rankings despite training on multiple Romance languages. This is to be expected, as languages in the same family share the same root and possess a big overlap in vocabulary. What is more interesting is that the classifiers trained on unrelated families, for example Nordic (MKC+): $\rho_s = 0.88$, have nearly matching Romance correlation despite limited lexical and syntactic overlap. Finally, we can see that removing French from Romance hurts more than expected. Romance without French (MKC-e) drops to $\rho_s = 0.71$, correlating worse than some linguistically distant families like Germanic or Uralic. This suggests potential syntactic interference: when the classifier trains on closely related but distinct languages (Spanish, Italian, Portuguese), it may learn Romance-specific syntactic patterns that don't perfectly generalize to French, obscuring the underlying quality signal.

\begin{figure}[t]
    \centering
    \includegraphics[width=0.8\textwidth]{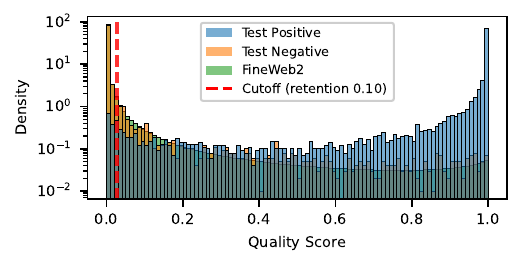}
    \caption{French quality score distribution from a Nordic classifier. The classifier effectively separates high-quality (positives) from low-quality (negatives) French text despite being trained solely on Nordic languages, suggesting the practical effectiveness of cross-lingual quality transfer.}
    \label{fig:score_distributions_nordic}
\end{figure}

To investigate this further, we plot the score distribution of the Nordic classifier in Figure~\ref{fig:score_distributions_nordic} to give us insights on how the classifier behaves. We apply it on a held-out set of French high-quality samples, negative samples, and the general FineWeb2. As we can observe, the classifier recognizes almost perfectly the negative and positive classes.  It applies a very strict threshold (90th percentile: 0.027) and assigns generally lower scores, yet still identifies high-quality content effectively. Additional experiments, including the baseline plot, can be found in the Appendix~\ref{appendix:score_distribution}. 

Additionally, we train our 1B models with the datasets from our classifiers. Despite the high rank correlation, we would expect the classifier trained on Romance to perform better than Nordic, even though both have never seen French samples in their training data. And we expect the Romance classifier to perform close to the HQ baseline, as these languages should be similar enough to French to give the classifier a good idea of what a high-quality sample should be. 

We show the results of the experiment in Table~\ref{tab:nordicromancebenchmark}, which provides a surprising result. The Romance classifier has the same rank as the HQ baseline, however, with slightly less in normalized accuracy. Furthermore, the Nordic classifier is the one with the highest mean normalized accuracy, which even outperformed the HQ baseline trained on French. 

\begin{table}[t]

\caption{Models trained on French data filtered by "Nordic" and "Romance (no French)" classifiers. The Nordic classifier outperforms the native French HQ baseline in aggregate accuracy, providing evidence for transferable quality signals across language families.}
\label{tab:nordicromancebenchmark}

\begin{center}
\begin{tabular}{lcccc}
\multicolumn{1}{c}{\bf Benchmark} &
\multicolumn{1}{c}{\bf No filtering} &
\multicolumn{1}{c}{\bf HQ} &
\multicolumn{1}{c}{\bf Romance (no fra)} &
\multicolumn{1}{c}{\bf Nordic}\\
\midrule
ARC-Challenge & 0.2891 & 0.3071 & 0.3054 & \textbf{0.3157} \\
Belebele\_c   & 0.3444 & 0.3511 & \textbf{0.3689} & 0.3422 \\
GMMLU\_c      & 0.2625 & 0.2925 & 0.2750 & \textbf{0.3075} \\
HellaSwag     & 0.4883 & 0.4748 & \textbf{0.4908} & 0.4673 \\
Include\_c    & 0.3866 & 0.4153 & 0.4272 & \textbf{0.4535} \\
M\_MMLU\_c    & 0.2831 & 0.2945 & \textbf{0.2950} & 0.2936 \\
XNLI          & 0.4707 & \textbf{0.4855} & 0.4827 & 0.4783 \\
XWinograd     & 0.6386 & 0.6506 & 0.6024 & \textbf{0.6627} \\
\hline
\textbf{Aggregate acc\_norm} & 0.3954 & 0.4089 & 0.4059 & \textbf{0.4151} \\
\textbf{Average rank} & 3.50 & \textbf{2.12} & \textbf{2.12} & 2.25 \\
\hline
\end{tabular}
\end{center}

\end{table}

The success of the Nordic classifier provides evidence for transferable quality features across language families. Because Nordic languages (Swedish, Danish, Norwegian, Icelandic) share no lexical or syntactic overlap with French, the classifier cannot rely on surface-level patterns.

These results suggest that cross-lingual transfer of quality signals is practically effective even across distant language families, though whether this reflects abstract semantic structure or shared formatting regularities in the embedding space remains to be characterized. However, we note that these high rank correlations may partially reflect the classifier learning shared dataset artifacts such as Wikipedia formatting or instruction-tuning templates rather than a purely abstract representation of quality. 
Nonetheless, capturing these cross-lingual artifacts serves as a highly effective and robust proxy for pretraining data selection.
Conversely, the underperformance of Romance-without-French suggests that syntactic similarity can be a double-edged sword: the classifier may overfit to Spanish/Italian grammatical structures that don't perfectly align with French, creating systematic biases that distant language families avoid.
\paragraph{Refining the Decision Boundary.}

We have observed that training on multiple languages provides strong performance boosts, and also can generalize to unseen samples. Nevertheless, upon examining the training pipeline, one could argue that using random samples from FineWeb2 might not be the best strategy for selecting negative samples.
Standard negative sampling draws randomly from FineWeb2, teaching the classifier to distinguish quality content from typical web noise (advertisements, navigation menus, broken formatting, spam). However, as filtering becomes more precise, a subtler distinction emerges: \textit{separating high-quality content from fluent but low-utility text}.

The Q3 strategy addresses this by sampling negatives from the 50th-75th percentile of the baseline HQ distribution. These samples are grammatically correct and well-formatted, but lacking in information density, logical depth, or educational value. We provide an example in the Appendix~\ref{appendix:samplesq3}.

In this approach, once a classifier is obtained, we apply it on FineWeb2 and extract samples from the third quartile (i.e. 50th to 75th percentile) as our negative samples for the training of a new classifier. We conduct these experiments using the HQ baseline (Q3) and the ML classifier (ML Q3). The ML classifier uses negative samples from the third quartile of its scores. We apply this extra step to all languages that have more than 200,000 samples in FineWeb2 to ensure we will not need aggressive token replication. 

\begin{table}[t]

\caption{Comparison of benchmark results using standard random negatives vs. Q3 negatives (fluent but low-utility text) in French. Q3 sampling consistently yields superior aggregate performance, with the monolingual classifier (Q3) performing slightly better than the ML (Q3) classifier.}
\label{benchmark:frenchq3}

\begin{center}
\begin{tabular}{lccccc}
\multicolumn{1}{c}{\bf Benchmark} &
\multicolumn{1}{c}{\bf No filtering} &
\multicolumn{1}{c}{\bf HQ} &
\multicolumn{1}{c}{\bf Q3} &
\multicolumn{1}{c}{\bf ML} &
\multicolumn{1}{c}{\bf ML (Q3)}\\
\midrule
ARC-Challenge & 0.2891 & 0.3071 & 0.3105 & \textbf{0.3165} & 0.3054 \\
Belebele\_c   & 0.3444 & 0.3511 & \textbf{0.3711} & 0.3511 & 0.3633 \\
GMMLU\_c      & 0.2625 & 0.2925 & 0.3025 & 0.2900 & \textbf{0.3075} \\
HellaSwag     & 0.4883 & 0.4748 & 0.4727 & \textbf{0.4956} & 0.4911 \\
Include\_c    & 0.3866 & 0.4153 & 0.4272 & 0.4224 & \textbf{0.4582} \\
M\_MMLU\_c    & 0.2831 & 0.2945 & 0.2992 & 0.2927 & \textbf{0.3007} \\
XNLI          & 0.4707 & 0.4855 & \textbf{0.4876} & 0.4807 & 0.4767 \\
XWinograd     & 0.6386 & 0.6506 & \textbf{0.6627} & 0.6024 & 0.6145 \\
\hline
\textbf{Aggregate acc\_norm} & 0.3954 & 0.4089 & \textbf{0.4167} & 0.4064 & 0.4147 \\
\textbf{Average rank} & 4.50 & 3.00 & \textbf{2.00} & 3.00 & 2.38 \\
\hline
\end{tabular}
\end{center}

\end{table}

\begin{table}[t]

\caption{Comparison of benchmark results using standard random negatives vs. Q3 negatives (fluent but low-utility text) in Spanish. Q3 sampling consistently yields superior aggregate performance.}
\label{benchmark:spanishq3}

\begin{center}
\begin{tabular}{lcccc}
\multicolumn{1}{c}{\bf Benchmark} &
\multicolumn{1}{c}{\bf No filtering} &
\multicolumn{1}{c}{\bf HQ} &
\multicolumn{1}{c}{\bf ML} &
\multicolumn{1}{c}{\bf ML (Q3)}\\
\midrule
ARC-Challenge & 0.2991 & 0.3077 & 0.3248 & \textbf{0.3282} \\
Belebele\_c   & 0.3422 & \textbf{0.3533} & 0.3456 & \textbf{0.3533} \\
GMMLU\_c      & 0.3100 & 0.3150 & 0.3250 & \textbf{0.3275} \\
HellaSwag     & 0.5006 & 0.5263 & 0.5310 & \textbf{0.5372} \\
Include\_c    & 0.3491 & 0.3727 & 0.3891 & \textbf{0.4000} \\
M\_MMLU\_c    & 0.2814 & 0.2985 & 0.3050 & \textbf{0.3083} \\
XNLI          & 0.4618 & 0.4538 & \textbf{0.4783} & 0.4667 \\
\hline
\textbf{Aggregate acc\_norm} & 0.3635 & 0.3753 & 0.3855 & \textbf{0.3887} \\
\textbf{Average rank} & 3.86 & 2.86 & 2.00 & \textbf{1.14} \\
\hline
\end{tabular}
\end{center}

\end{table}

Tables~\ref{benchmark:frenchq3} and~\ref{benchmark:spanishq3} evaluate the impact of refining the negative sampling strategy using third-quartile bootstrapping (Q3). Compared to random negatives from FineWeb2, harder negatives consistently improve performance, confirming that sharper decision boundaries emerge when the classifier is trained on more ambiguous examples.

For French, the monolingual Q3 classifier achieves the strongest aggregate performance, while the multilingual bootstrapped model (ML Q3) remains highly competitive, with only a marginal drop in normalized accuracy. In contrast, for Spanish, ML Q3 yields the best overall performance across benchmarks, surpassing both standard multilingual filtering and monolingual HQ.

Taken together, these results indicate that bootstrapped negative sampling systematically strengthens quality discrimination. While language-specific refinement can yield peak in-language performance, the multilingual Q3 model matches or exceeds these gains in some languages and, critically, preserves cross-lingual transfer without retraining. This further supports the existence of shared  in representation space that can be progressively refined through Q3 sampling.

\paragraph{Tuning the Retention Rate.}
While FineWeb2-HQ was derived using a retention rate of 10\% for high-resource languages, we argue that this hyperparameter should be tuned for each language separately. The rate for which we filter samples for a language can have a big impact on the filtered data.
\begin{table}[t]
\caption{Comparison between 10\% and 15\% retention using the ML classifier. Increasing the retention rate to 15\% improves accuracy across most benchmarks, suggesting the default 10\% threshold is overly aggressive when using our classifier on French.}
\label{tab:french}
\begin{center}
\begin{tabular}{lcc}
\multicolumn{1}{c}{\bf Benchmark} &
\multicolumn{1}{c}{\bf ML} &
\multicolumn{1}{c}{\bf ML (15\%)}\\
\midrule
ARC-Challenge & \textbf{0.3165} & 0.3139 \\
Belebele\_c & 0.3511 & \textbf{0.3611} \\
GMMLU\_c & \textbf{0.2900} & 0.2875 \\
HellaSwag & 0.4956 & \textbf{0.5135} \\
Include\_c & 0.4224 & \textbf{0.4726} \\
M\_MMLU\_c & 0.2927 & \textbf{0.2991} \\
XNLI & \textbf{0.4807} & \textbf{0.4807} \\
XWinograd & 0.6024 & \textbf{0.6386} \\
\hline
\textbf{Aggregate acc\_norm} & 0.4064 & \textbf{0.4209} \\
\textbf{Average rank} & 1.62 & \textbf{1.25} \\
\hline
\end{tabular}
\end{center}
\end{table}

\begin{table}[t]
\caption{Comparison between 10\% and 15\% retention using the ML classifier on Spanish. Similar to French, the 15\% retention rate yields a higher aggregate accuracy and better average rank when using our ML classifier.}
\label{tab:spanish}
\begin{center}
\begin{tabular}{lcc}
\multicolumn{1}{c}{\bf Benchmark} &
\multicolumn{1}{c}{\bf ML} &
\multicolumn{1}{c}{\bf ML (15\%)}\\
\midrule
ARC-Challenge & 0.3248 & \textbf{0.3436} \\
Belebele\_c & 0.3456 & \textbf{0.3500} \\
GMMLU\_c & \textbf{0.3250} & 0.3225 \\
HellaSwag & 0.5310 & \textbf{0.5440} \\
Include\_c & 0.3891 & \textbf{0.4036} \\
M\_MMLU\_c & 0.3050 & \textbf{0.3080} \\
XNLI & \textbf{0.4783} & 0.4562 \\
\hline
\textbf{Aggregate acc\_norm} & 0.3855 & \textbf{0.3897} \\
\textbf{Average rank} & 1.71 & \textbf{1.29} \\
\hline
\end{tabular}
\end{center}
\end{table}

\begin{table}[t]
\caption{Evaluation of standard (56\%) vs. aggressive (10\%, 20\%) filtering with replication. Unlike high-resource languages, reducing Arabic retention to increase token replication degrades performance.}
\label{tab:arabic}
\begin{center}
\begin{tabular}{lcccccc}
\multicolumn{1}{c}{\bf Benchmark} &
\multicolumn{1}{c}{\bf HQ} &
\multicolumn{1}{c}{\bf ML} &
\multicolumn{1}{c}{\bf HQ (10\%)} &
\multicolumn{1}{c}{\bf HQ (20\%)} &
\multicolumn{1}{c}{\bf ML (10\%)} &
\multicolumn{1}{c}{\bf ML (20\%)}\\
\midrule
ARC-Easy & \textbf{0.2898} & 0.2855 & 0.2720 & 0.2771 & 0.2741 & 0.2762 \\
AlGhafa PIQA-MT & \textbf{0.5145} & 0.5112 & 0.5019 & 0.5090 & 0.4948 & 0.5128 \\
AlGhafa RACE & 0.2775 & \textbf{0.2883} & 0.2747 & 0.2834 & 0.2765 & 0.2786 \\
AlGhafa SciQ & 0.4432 & 0.4503 & 0.4503 & 0.4171 & \textbf{0.4573} & 0.4563 \\
ARC-Challenge & 0.2660 & \textbf{0.2797} & 0.2669 & 0.2772 & 0.2712 & 0.2626 \\
Belebele\_c & 0.3233 & 0.3122 & \textbf{0.3378} & 0.2944 & 0.3278 & 0.3256 \\
GMMLU\_c & 0.2575 & 0.2700 & 0.2475 & 0.2550 & \textbf{0.2725} & 0.2525 \\
HellaSwag & 0.3873 & \textbf{0.3925} & 0.3592 & 0.3728 & 0.3747 & 0.3857 \\
Include\_c & 0.2681 & \textbf{0.3043} & 0.2717 & 0.2772 & 0.2844 & 0.2790 \\
M\_MMLU\_c & 0.2614 & 0.2646 & 0.2652 & \textbf{0.2674} & 0.2618 & 0.2634 \\
AlGhafa PIQA & 0.6088 & 0.6110 & 0.5958 & 0.6023 & 0.6039 & \textbf{0.6143} \\
XNLI & 0.3309 & 0.3349 & 0.3325 & 0.3333 & 0.3325 & \textbf{0.3357} \\
XStoryCloze & \textbf{0.5923} & 0.5811 & 0.5877 & 0.5817 & 0.5784 & 0.5804 \\
\hline
\textbf{Aggregate acc\_norm} & 0.3708 & \textbf{0.3758} & 0.3664 & 0.3652 & 0.3700 & 0.3710 \\
\textbf{Average rank} & 3.62 & \textbf{2.31} & 4.31 & 3.69 & 3.69 & 3.23 \\
\hline
\end{tabular}
\end{center}
\end{table}

Tables~\ref{tab:french}, \ref{tab:spanish}, and \ref{tab:arabic} analyze the effect of tuning the retention rate used during multilingual and monolingual filtering. Increasing the retention rate consistently improves aggregate performance in French and Spanish, with ML at 15\% outperforming the default 10\% across most benchmarks. Gains are particularly pronounced on knowledge-intensive tasks such as Include~\citep{include} and HellaSwag~\citep{zellers2019hellaswag}, indicating that overly aggressive filtering can discard useful high-quality content.

In Arabic, instead of using 56\% like for FineWeb2-HQ, we try a retention rate of 10\% and 20\%, resulting in a replication of tokens of 10x and 5x, respectively. We see that the standard retention of 56\% yields the best overall performance when combined with multilingual filtering. This suggests that optimal retention rates are language-dependent, and that aggressive repetition of the same high-quality data does not necessarily lead to the best performance. We further note that the ML gain over HQ for Arabic (~0.5\%) is comparable to the seed variance reported in Appendix H (~0.3\%), and should therefore be interpreted with caution pending multi-seed validation.

\begin{table}[t]
\caption{Aggregated macro and micro metrics for all strategies using a 10\% retention rate. The ML (Q3) strategy achieves the best overall performance in both rank and normalized accuracy, highlighting the robustness of combining both our methods to obtain a robust general multilingual classifier.}
\label{tab:global_ranks}
\begin{center}
\begin{tabular}{lcccc}
\multicolumn{1}{c}{\bf Method} &
\multicolumn{1}{c}{\bf Macro Rank} &
\multicolumn{1}{c}{\bf Micro Rank} &
\multicolumn{1}{c}{\bf Macro Acc} &
\multicolumn{1}{c}{\bf Micro Acc}\\
\midrule
ML (Q3) & \textbf{1.6983} & \textbf{1.7857} & \textbf{0.4082} & \textbf{0.4113} \\
ML & 1.9808 & 1.9286 & 0.4052 & 0.4092 \\
HQ & 2.4556 & 2.4286 & 0.4011 & 0.4052 \\
No filtering & 3.7248 & 3.7143 & 0.3871 & 0.3907 \\
\hline
\end{tabular}
\end{center}
\end{table}

\paragraph{Synthesis: When Does Multilingual Pooling Help?}

Across our four languages, we observe a pattern: the key insight is not that multilingual pooling \textit{always} dominates, but that it provides a reliable baseline across languages, while language-specific optimization (negative sampling strategy, retention rate, anchor curation) can yield comparable or superior results when tuned appropriately. These findings provide evidence that the classifier learns language-agnostic markers in the embedding space by training on multiple languages. 
Note on Stochasticity: While multilingual pooling improves overall rank stability, our ablations show that downstream aggregate accuracy remains sensitive to the classifier's initial sampling seed (shifting by ~0.3\% in Arabic and ~0.8\% in French). We provide a more detailed analysis of these seed variances in Appendix~\ref{appendix:seed}.

\section{Conclusion}

This work investigates whether quality classifiers can generalize across languages by exploiting shared semantic structures in multilingual embedding spaces. Through systematic evaluation across French, Spanish, Arabic, and Chinese, we demonstrate that classifiers trained on typologically distant language families can effectively filter quality content in unrelated languages.
Across the four tested languages, multilingual classifiers improved over monolingual baselines, with particularly strong gains in Spanish (3.0 ranks) and Arabic (2.31 ranks). We also introduce the Q3 sampling strategy, which refines decision boundaries by training against fluent but low-utility text rather than random negatives, offering a complementary approach to multilingual pooling for high-resource languages.
No single strategy dominates. For French, Q3 negatives, ML (15\%), and ML (Q3) all achieve comparable results, suggesting practitioners should select based on computational constraints and available data. The 10\% threshold from prior work may be overly aggressive when using our ML classifier; increasing to 15\% improved French performance substantially (40.64\% to 42.09\%), which indicates that this hyperparameter needs language-specific tuning. However, when comparing all approaches (Table~\ref{tab:global_ranks}), we find that the ML (Q3) strategy is the one dominating in terms of performance, which means combining our methods leads to better results. 
Overall, our findings suggest that multilingual pooling can help democratize high-quality data curation for underrepresented languages while boosting performance on high-resource languages, though the effectiveness varies by language family and resource level.

\bibliography{iclr2026_conference}
\bibliographystyle{iclr2026_conference}

\appendix

\appendix

\section{Dataset Counts}
\label{sec:datacounts}

\setlength{\tabcolsep}{5pt}  
{
\scriptsize
\begin{longtable}{lrrrrrrrrrr}
\caption{Dataset Counts by Language (sorted by total count, descending). Note: OA = openassistant2, MMLU = openai\_mmlu, Inc = include, Tag = tagengo, Euro = euroblocks, Wiki = muri\_wikipedia, AH = aya\_human, AC = aya\_collection, WQA = wikiqa.} \label{tab:dataset_counts} \\
\multicolumn{1}{c}{\bf Language} &
\multicolumn{1}{c}{\bf OA} &
\multicolumn{1}{c}{\bf MMLU} &
\multicolumn{1}{c}{\bf Inc} &
\multicolumn{1}{c}{\bf Tag} &
\multicolumn{1}{c}{\bf Euro} &
\multicolumn{1}{c}{\bf Wiki} &
\multicolumn{1}{c}{\bf AH} &
\multicolumn{1}{c}{\bf AC} &
\multicolumn{1}{c}{\bf WQA} &
\multicolumn{1}{c}{\bf Total}\\
\midrule
\endfirsthead
\multicolumn{11}{c}{\tablename\ \thetable\ -- Continued from previous page} \\
 \\
\multicolumn{1}{c}{\bf Language} &
\multicolumn{1}{c}{\bf OA} &
\multicolumn{1}{c}{\bf MMLU} &
\multicolumn{1}{c}{\bf Inc} &
\multicolumn{1}{c}{\bf Tag} &
\multicolumn{1}{c}{\bf Euro} &
\multicolumn{1}{c}{\bf Wiki} &
\multicolumn{1}{c}{\bf AH} &
\multicolumn{1}{c}{\bf AC} &
\multicolumn{1}{c}{\bf WQA} &
\multicolumn{1}{c}{\bf Total}\\
\midrule
\endhead

\hline
\multicolumn{11}{r}{Continued on next page} \\
\endfoot

\hline
\endlastfoot

eng\_Latn & 61,278 & 99,842 & 0 & 15,771 & 151,135 & 6,657 & 3,944 & 14,693,823 & 0 & 15,032,450 \\
jpn\_Jpan & 756 & 14,042 & 501 & 2,521 & 4,735 & 6,971 & 6,259 & 6,218,459 & 0 & 6,254,244 \\
arb\_Arab & 76 & 14,042 & 552 & 789 & 0 & 8,508 & 4,995 & 5,857,458 & 17,706 & 5,904,126 \\
tha\_Thai & 1,503 & 0 & 0 & 133 & 0 & 7,187 & 724 & 5,338,232 & 0 & 5,347,779 \\
deu\_Latn & 5,797 & 14,042 & 139 & 5,739 & 14,081 & 7,009 & 241 & 4,689,989 & 0 & 4,737,037 \\
fra\_Latn & 3,686 & 14,042 & 419 & 5,369 & 14,882 & 6,988 & 1,422 & 4,285,094 & 0 & 4,331,902 \\
tel\_Telu & 0 & 0 & 548 & 0 & 0 & 7,765 & 8,439 & 4,058,535 & 0 & 4,075,287 \\
rus\_Cyrl & 13,336 & 0 & 552 & 8,056 & 4,727 & 7,042 & 423 & 4,005,166 & 0 & 4,039,302 \\
fin\_Latn & 138 & 0 & 551 & 92 & 1,022 & 6,970 & 742 & 3,939,941 & 16,383 & 3,965,839 \\
spa\_Latn & 26,811 & 14,042 & 550 & 8,318 & 17,428 & 7,012 & 3,854 & 3,872,864 & 0 & 3,950,879 \\
ita\_Latn & 899 & 14,042 & 548 & 7,063 & 15,963 & 7,100 & 738 & 3,890,852 & 0 & 3,937,205 \\
urd\_Arab & 0 & 0 & 352 & 3 & 0 & 7,893 & 654 & 3,876,197 & 0 & 3,885,099 \\
pol\_Latn & 431 & 0 & 548 & 1,090 & 5,358 & 7,013 & 1,483 & 3,841,451 & 16,964 & 3,874,338 \\
por\_Latn & 2,581 & 14,042 & 551 & 12,564 & 13,966 & 7,367 & 8,997 & 3,786,062 & 0 & 3,846,130 \\
hin\_Deva & 0 & 14,042 & 547 & 20 & 7,982 & 7,217 & 1,153 & 3,772,864 & 0 & 3,803,825 \\
fas\_Arab & 0 & 0 & 548 & 184 & 0 & 7,504 & 1,578 & 3,785,250 & 5,615 & 3,800,679 \\
nld\_Latn & 72 & 0 & 551 & 383 & 7,683 & 6,840 & 1,733 & 3,736,938 & 18,723 & 3,772,923 \\
ukr\_Cyrl & 821 & 0 & 550 & 323 & 5,191 & 0 & 522 & 3,729,748 & 12,979 & 3,750,134 \\
ces\_Latn & 12 & 0 & 0 & 179 & 4,105 & 6,793 & 0 & 3,719,214 & 11,203 & 3,741,506 \\
heb\_Hebr & 24 & 0 & 550 & 120 & 0 & 6,916 & 0 & 3,658,066 & 17,229 & 3,682,905 \\
cmn\_Hani & 0 & 14,042 & 545 & 5,338 & 27,507 & 9,368 & 4,909 & 3,606,935 & 0 & 3,668,644 \\
hun\_Latn & 113 & 0 & 550 & 214 & 4,010 & 7,096 & 98 & 3,637,911 & 14,320 & 3,664,312 \\
swe\_Latn & 1 & 0 & 0 & 256 & 6,476 & 6,524 & 1,310 & 3,632,622 & 14,650 & 3,661,839 \\
tur\_Latn & 37 & 0 & 548 & 406 & 0 & 7,084 & 4,046 & 3,628,109 & 18,422 & 3,658,652 \\
kor\_Hang & 20 & 14,042 & 500 & 1,609 & 2,905 & 7,343 & 361 & 3,605,894 & 17,616 & 3,650,290 \\
cat\_Latn & 1,194 & 0 & 0 & 73 & 223 & 7,209 & 0 & 3,625,537 & 15,438 & 3,649,674 \\
srp\_Cyrl & 0 & 0 & 550 & 6 & 0 & 7,529 & 152 & 3,636,573 & 0 & 3,644,810 \\
ben\_Beng & 1 & 14,042 & 548 & 0 & 0 & 7,609 & 1,534 & 3,601,287 & 8,532 & 3,633,553 \\
vie\_Latn & 203 & 0 & 550 & 429 & 0 & 7,040 & 8,676 & 3,613,270 & 0 & 3,630,168 \\
ind\_Latn & 12 & 14,042 & 550 & 240 & 0 & 0 & 786 & 3,610,078 & 0 & 3,625,708 \\
ron\_Latn & 0 & 0 & 0 & 71 & 3,993 & 7,170 & 0 & 3,602,212 & 11,188 & 3,624,634 \\
bul\_Cyrl & 0 & 0 & 550 & 56 & 210 & 7,221 & 0 & 3,602,878 & 11,207 & 3,622,122 \\
hau\_Latn & 0 & 0 & 0 & 0 & 0 & 8,045 & 3,512 & 3,608,883 & 0 & 3,620,440 \\
tam\_Taml & 0 & 0 & 550 & 5 & 0 & 7,766 & 14,133 & 3,596,707 & 0 & 3,619,161 \\
slk\_Latn & 0 & 0 & 0 & 17 & 990 & 7,061 & 0 & 3,594,203 & 15,036 & 3,617,307 \\
slv\_Latn & 0 & 0 & 0 & 10 & 201 & 6,873 & 0 & 3,593,626 & 16,125 & 3,616,835 \\
dan\_Latn & 40 & 0 & 0 & 67 & 4 & 6,348 & 97 & 3,601,900 & 8,212 & 3,616,668 \\
ell\_Grek & 0 & 0 & 552 & 308 & 582 & 7,495 & 623 & 3,606,249 & 827 & 3,616,636 \\
yor\_Latn & 0 & 14,042 & 0 & 0 & 0 & 0 & 11,758 & 3,587,233 & 0 & 3,613,033 \\
zsm\_Latn & 0 & 0 & 501 & 5 & 0 & 8,060 & 10,073 & 3,593,313 & 0 & 3,611,952 \\
bel\_Cyrl & 0 & 0 & 550 & 2 & 0 & 7,499 & 0 & 3,589,912 & 12,868 & 3,610,831 \\
sin\_Sinh & 0 & 0 & 0 & 5 & 0 & 7,290 & 14,524 & 3,587,051 & 0 & 3,608,870 \\
plt\_Latn & 0 & 0 & 0 & 0 & 0 & 6,895 & 14,597 & 3,586,962 & 0 & 3,608,454 \\
ibo\_Latn & 0 & 0 & 0 & 0 & 0 & 8,767 & 1,534 & 3,597,292 & 0 & 3,607,593 \\
swh\_Latn & 0 & 14,042 & 0 & 0 & 0 & 7,518 & 366 & 3,580,061 & 0 & 3,601,987 \\
ary\_Arab & 0 & 0 & 0 & 0 & 0 & 0 & 8,090 & 3,591,621 & 0 & 3,599,711 \\
glg\_Latn & 0 & 0 & 0 & 0 & 0 & 6,906 & 0 & 3,572,365 & 19,481 & 3,598,752 \\
lit\_Latn & 0 & 0 & 534 & 11 & 0 & 7,239 & 916 & 3,573,281 & 16,354 & 3,598,335 \\
amh\_Ethi & 0 & 0 & 0 & 3 & 0 & 7,132 & 1,207 & 3,589,993 & 0 & 3,598,335 \\
nob\_Latn & 0 & 0 & 0 & 26 & 534 & 6,870 & 0 & 3,572,365 & 17,742 & 3,597,537 \\
eus\_Latn & 257 & 0 & 500 & 6 & 0 & 7,075 & 939 & 3,573,304 & 15,069 & 3,597,150 \\
ltz\_Latn & 0 & 0 & 0 & 1 & 0 & 6,999 & 0 & 3,572,365 & 17,689 & 3,597,054 \\
som\_Latn & 0 & 0 & 0 & 0 & 0 & 7,036 & 7,704 & 3,582,111 & 0 & 3,596,851 \\
ekk\_Latn & 0 & 0 & 224 & 18 & 179 & 7,028 & 0 & 3,572,365 & 16,824 & 3,596,638 \\
isl\_Latn & 0 & 0 & 0 & 5 & 18 & 7,372 & 0 & 3,572,365 & 15,426 & 3,595,186 \\
gla\_Latn & 0 & 0 & 0 & 0 & 0 & 7,655 & 0 & 3,572,365 & 15,053 & 3,595,073 \\
mkd\_Cyrl & 0 & 0 & 551 & 5 & 0 & 7,548 & 0 & 3,572,365 & 14,066 & 3,594,535 \\
lvs\_Latn & 0 & 0 & 0 & 22 & 176 & 7,515 & 0 & 3,572,365 & 14,311 & 3,594,389 \\
als\_Latn & 0 & 0 & 551 & 5 & 0 & 8,152 & 120 & 3,572,485 & 12,705 & 3,594,018 \\
ydd\_Hebr & 0 & 0 & 0 & 2 & 0 & 7,062 & 0 & 3,572,365 & 13,417 & 3,592,846 \\
mlt\_Latn & 0 & 0 & 0 & 0 & 0 & 4,771 & 0 & 3,572,365 & 15,309 & 3,592,445 \\
mar\_Deva & 0 & 0 & 0 & 1 & 0 & 7,743 & 3,545 & 3,579,228 & 0 & 3,590,517 \\
cym\_Latn & 0 & 0 & 0 & 0 & 0 & 6,944 & 0 & 3,572,365 & 11,044 & 3,590,353 \\
guj\_Gujr & 0 & 0 & 0 & 0 & 0 & 7,492 & 3,989 & 3,578,511 & 0 & 3,589,992 \\
mal\_Mlym & 0 & 0 & 479 & 2 & 0 & 7,660 & 1,749 & 3,577,960 & 0 & 3,587,850 \\
nno\_Latn & 0 & 0 & 0 & 0 & 0 & 0 & 0 & 3,572,365 & 14,518 & 3,586,883 \\
npi\_Deva & 0 & 0 & 500 & 0 & 0 & 5,020 & 4,002 & 3,576,367 & 0 & 3,585,889 \\
sna\_Latn & 0 & 0 & 0 & 0 & 0 & 7,457 & 1,368 & 3,576,309 & 0 & 3,585,134 \\
zul\_Latn & 0 & 0 & 0 & 0 & 0 & 7,642 & 1,833 & 3,574,437 & 0 & 3,583,912 \\
afr\_Latn & 0 & 0 & 0 & 2 & 0 & 6,503 & 0 & 3,577,285 & 0 & 3,583,790 \\
kan\_Knda & 0 & 0 & 0 & 1 & 0 & 7,574 & 334 & 3,573,855 & 0 & 3,581,764 \\
gle\_Latn & 0 & 0 & 0 & 0 & 0 & 6,549 & 1,245 & 3,573,610 & 0 & 3,581,404 \\
ceb\_Latn & 0 & 0 & 0 & 0 & 0 & 7,130 & 727 & 3,573,092 & 0 & 3,580,949 \\
mya\_Mymr & 0 & 0 & 0 & 2 & 0 & 7,367 & 472 & 3,572,837 & 0 & 3,580,678 \\
hat\_Latn & 0 & 0 & 0 & 0 & 0 & 7,982 & 106 & 3,572,471 & 0 & 3,580,559 \\
kaz\_Cyrl & 0 & 0 & 500 & 2 & 0 & 7,547 & 0 & 3,572,365 & 0 & 3,580,414 \\
snd\_Arab & 0 & 0 & 0 & 0 & 0 & 7,470 & 274 & 3,572,639 & 0 & 3,580,383 \\
azj\_Latn & 0 & 0 & 548 & 4 & 0 & 7,313 & 0 & 3,572,365 & 0 & 3,580,230 \\
kat\_Geor & 0 & 0 & 500 & 1 & 0 & 7,351 & 0 & 3,572,365 & 0 & 3,580,217 \\
jav\_Latn & 0 & 0 & 0 & 0 & 0 & 6,421 & 247 & 3,573,441 & 0 & 3,580,109 \\
khm\_Khmr & 0 & 0 & 0 & 1 & 0 & 7,714 & 0 & 3,572,365 & 0 & 3,580,080 \\
epo\_Latn & 269 & 0 & 0 & 17 & 0 & 7,266 & 0 & 3,572,365 & 0 & 3,579,917 \\
khk\_Cyrl & 0 & 0 & 0 & 6 & 0 & 7,199 & 0 & 3,572,365 & 0 & 3,579,570 \\
hye\_Armn & 0 & 0 & 550 & 3 & 0 & 0 & 0 & 3,576,382 & 0 & 3,576,935 \\
xho\_Latn & 0 & 0 & 0 & 0 & 0 & 1,351 & 377 & 3,574,806 & 0 & 3,576,534 \\
lao\_Laoo & 0 & 0 & 0 & 1 & 0 & 3,672 & 0 & 3,572,365 & 0 & 3,576,038 \\
pbt\_Arab & 0 & 0 & 0 & 0 & 0 & 0 & 989 & 3,573,354 & 0 & 3,574,343 \\
sun\_Latn & 0 & 0 & 0 & 0 & 0 & 90 & 194 & 3,573,767 & 0 & 3,574,051 \\
arz\_Arab & 0 & 0 & 0 & 0 & 0 & 0 & 529 & 3,572,894 & 0 & 3,573,423 \\
sot\_Latn & 0 & 0 & 0 & 0 & 0 & 658 & 0 & 3,572,365 & 0 & 3,573,023 \\
ars\_Arab & 0 & 0 & 0 & 0 & 0 & 0 & 136 & 3,572,501 & 0 & 3,572,637 \\
apc\_Arab & 0 & 0 & 0 & 0 & 0 & 0 & 81 & 3,572,446 & 0 & 3,572,527 \\
ckb\_Arab & 0 & 0 & 0 & 0 & 0 & 0 & 79 & 3,572,444 & 0 & 3,572,523 \\
lat\_Latn & 0 & 0 & 0 & 4 & 0 & 7,756 & 0 & 0 & 21,836 & 29,596 \\
lij\_Latn & 0 & 0 & 0 & 0 & 0 & 6,715 & 0 & 5,955 & 16,162 & 28,832 \\
oci\_Latn & 0 & 0 & 0 & 0 & 0 & 6,813 & 0 & 0 & 17,194 & 24,007 \\
lim\_Latn & 0 & 0 & 0 & 0 & 0 & 6,693 & 0 & 0 & 16,668 & 23,361 \\
nds\_Latn & 0 & 0 & 0 & 0 & 0 & 6,988 & 0 & 0 & 16,216 & 23,204 \\
vec\_Latn & 0 & 0 & 0 & 0 & 0 & 5,424 & 0 & 0 & 17,749 & 23,173 \\
scn\_Latn & 0 & 0 & 0 & 0 & 0 & 5,999 & 0 & 0 & 16,842 & 22,841 \\
pan\_Guru & 0 & 0 & 0 & 0 & 0 & 7,909 & 6,385 & 8,541 & 0 & 22,835 \\
bar\_Latn & 0 & 0 & 0 & 0 & 0 & 3,070 & 0 & 0 & 19,563 & 22,633 \\
hrv\_Latn & 0 & 0 & 550 & 11 & 30 & 0 & 0 & 6,913 & 14,966 & 22,470 \\
fao\_Latn & 0 & 0 & 0 & 0 & 0 & 6,065 & 0 & 0 & 16,102 & 22,167 \\
bre\_Latn & 0 & 0 & 0 & 1 & 0 & 6,381 & 0 & 0 & 15,530 & 21,912 \\
arg\_Latn & 0 & 0 & 0 & 0 & 0 & 5,304 & 0 & 0 & 15,439 & 20,743 \\
roh\_Latn & 0 & 0 & 0 & 0 & 0 & 2,062 & 0 & 0 & 17,823 & 19,885 \\
srd\_Latn & 0 & 0 & 0 & 0 & 0 & 5,606 & 0 & 0 & 13,850 & 19,456 \\
kmr\_Latn & 0 & 0 & 0 & 0 & 0 & 6,986 & 0 & 0 & 11,674 & 18,660 \\
ast\_Latn & 0 & 0 & 0 & 0 & 0 & 0 & 0 & 0 & 18,383 & 18,383 \\
fry\_Latn & 0 & 0 & 0 & 0 & 0 & 0 & 0 & 0 & 17,450 & 17,450 \\
bos\_Latn & 0 & 0 & 0 & 0 & 0 & 0 & 0 & 0 & 15,933 & 15,933 \\
sco\_Latn & 0 & 0 & 0 & 0 & 0 & 0 & 0 & 0 & 15,212 & 15,212 \\
dag\_Latn & 0 & 0 & 0 & 0 & 0 & 0 & 0 & 0 & 12,848 & 12,848 \\
szl\_Latn & 0 & 0 & 0 & 0 & 0 & 4,755 & 0 & 0 & 7,388 & 12,143 \\
fur\_Latn & 0 & 0 & 0 & 0 & 0 & 3,114 & 0 & 0 & 7,152 & 10,266 \\
lmo\_Latn & 0 & 0 & 0 & 0 & 0 & 0 & 0 & 0 & 9,153 & 9,153 \\
nap\_Latn & 0 & 0 & 0 & 0 & 0 & 0 & 0 & 0 & 7,879 & 7,879 \\
wol\_Latn & 0 & 0 & 0 & 0 & 0 & 857 & 2,914 & 3,146 & 0 & 6,917 \\
war\_Latn & 0 & 0 & 0 & 2 & 0 & 6,437 & 0 & 0 & 0 & 6,439 \\
pfl\_Latn & 0 & 0 & 0 & 0 & 0 & 0 & 0 & 0 & 6,321 & 6,321 \\
san\_Deva & 0 & 0 & 0 & 1 & 0 & 6,052 & 0 & 0 & 0 & 6,053 \\
tuk\_Latn & 0 & 0 & 0 & 1 & 0 & 5,919 & 0 & 0 & 0 & 5,920 \\
frp\_Latn & 0 & 0 & 0 & 0 & 0 & 0 & 0 & 0 & 5,123 & 5,123 \\
fil\_Latn & 0 & 0 & 0 & 0 & 0 & 0 & 1,241 & 1,241 & 0 & 2,482 \\
nya\_Latn & 0 & 0 & 0 & 0 & 0 & 853 & 688 & 688 & 0 & 2,229 \\
bod\_Tibt & 0 & 0 & 0 & 1 & 0 & 1,998 & 0 & 0 & 0 & 1,999 \\
ltg\_Latn & 0 & 0 & 0 & 0 & 0 & 0 & 0 & 0 & 1,872 & 1,872 \\
rmy\_Latn & 0 & 0 & 0 & 0 & 0 & 0 & 0 & 0 & 877 & 877 \\
anp\_Deva & 0 & 0 & 0 & 0 & 0 & 0 & 0 & 0 & 57 & 57 \\
\end{longtable}
}

\section{Score Distribution of Language Families classifier}
\label{appendix:score_distribution}
\begin{figure}[t]
    \centering
    \label{fig:score_distributions_romance}
    \includegraphics[width=0.8\textwidth]{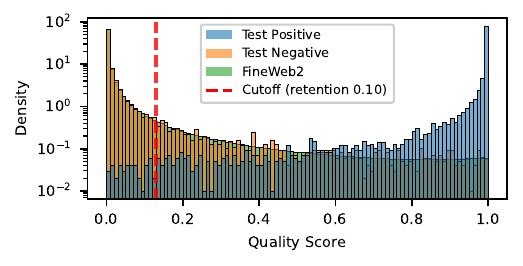}

    \vspace{0.4cm}

    \includegraphics[width=0.8\textwidth]{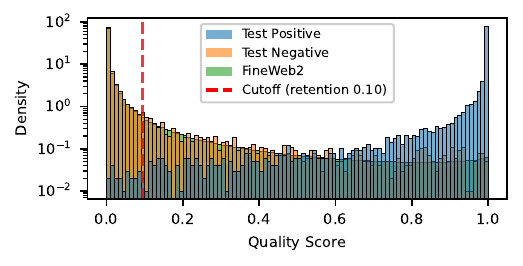}
    \caption{Distribution of quality scores for French baseline (top) and Romance languages classifier including French (bottom) on French FineWeb2 samples}

\end{figure}

\begin{figure}[t]
    \centering
    \label{fig:score_distribution_no_fra}
    \includegraphics[width=0.8\textwidth]{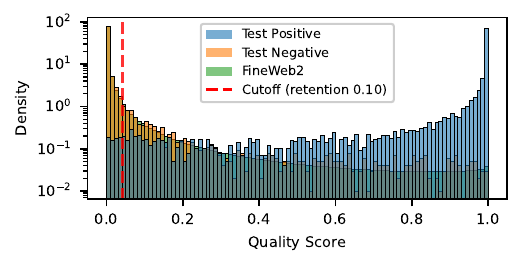}
     \caption{Distribution of quality scores for Romance languages classifier without French}

\end{figure}

Figures~\ref{fig:score_distributions_romance} and~\ref{fig:score_distributions_nordic} show the distribution of quality scores assigned to French documents by different classifiers. Several patterns emerge:

\begin{itemize}
    \item \textbf{French and Romance classifiers} (Fig~\ref{fig:score_distributions_romance}) exhibit similar bimodal distributions: most documents score near 0 (clear negatives) with a long tail toward 1 (clear positives). The 90th percentile cutoffs are comparable (French: ~0.045, Romance: ~0.048).
    
    \item \textbf{Nordic classifier} (Fig~\ref{fig:score_distributions_nordic}) produces a markedly different distribution despite successfully separating quality tiers. It applies a much stricter threshold (90th percentile: 0.027) and assigns generally lower scores, yet still identifies high-quality content effectively.
    
    \item \textbf{Romance without French} (Fig~\ref{fig:score_distribution_no_fra}) shows intermediate behavior, with more score mass in the middle range, suggesting less confident predictions.
\end{itemize}

\section{Qualitative Analysis of Classifier Rankings}
\label{appendix:samples}

This section provides a qualitative comparison of how different classifiers rank the same French documents. By observing extreme rank shifts, we can infer the ``features'' prioritized by each model.

\subsection{Language Family Classifiers Comparison on French Filtering}
\label{appendix:rank_corr}
We provide here the full table for all the language families classifiers filtering on French. 

\begin{table}[t]
\centering
\caption{Spearman and Kendall correlations between family-specific classifiers and the French HQ baseline. High correlation in distant families (e.g., Uralic, Nordic) suggests cross-lingual transfer of quality signals, while the drop in "Romance (no French)" suggests potential syntactic interference.}
\label{tab:fra_retention_correlations_appendix}
\begin{tabular}{lcc}
\toprule
\textbf{Experiment} & \textbf{Spearman} & \textbf{Kendall} \\
\midrule
Romance (spa, fra, por, ita, ron, cat) MKC+ & 0.8928 & 0.7173 \\
Uralic (fin, ekk, hun) MKC+ & 0.8870 & 0.7073 \\
Nordic (swe, dan, nob, isl) MKC+ & 0.8820 & 0.6990 \\
Nordic (swe, dan, nob, isl) MKC-e & 0.8790 & 0.7014 \\
Germanic (deu, nld, en, afr, ltz) MKC+ & 0.8750 & 0.6916 \\
Uralic (fin, ekk, hun) MKC-e & 0.8651 & 0.6823 \\
Germanic (deu, nld, en, afr, ltz) MKC-e & 0.8465 & 0.6585 \\
Slavic (pol, rus, ces, ukr, bul, srp, hrv) MKC+ & 0.8387 & 0.6484 \\
Romance (spa, fra, por, ita, ron, cat) MKC-e & 0.7808 & 0.5880 \\
Slavic (pol, rus, ces, ukr, bul, srp, hrv) MKC-e & 0.7443 & 0.5523 \\
Indo-Aryan (hin, urd, ben, pan, mar) MKC+ & 0.7157 & 0.5221 \\
Romance, no French (spa, por, ita, ron, cat) MKC-e & 0.7139 & 0.5228 \\
Indo-Aryan (hin, urd, ben, pan, mar) MKC-e & 0.6348 & 0.4565 \\
\bottomrule
\end{tabular}
\end{table}

\subsection{Monolingual (HQ) vs. Multilingual (ML) in French}

\begin{figure}[t]
    \centering
    \label{fig:retention_of_nonsense_text}
    \includegraphics[width=0.8\textwidth]{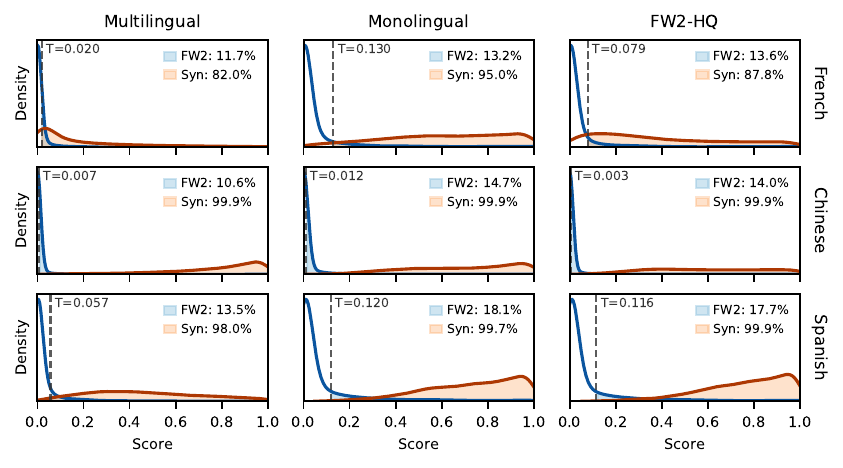}
    \caption{Comparison of score distributions for 10K synthetically generated grammatically correct nonsense samples across multilingual (\textit{ML}), monolingual (\textit{HQ}), and FineWeb2-HQ (\textit{FW2-HQ}) classifiers for French, Chinese, and Spanish, compared to 10K random samples from FineWeb2. Synthetic samples were generated using Qwen3-32B~\citep{qwen3technicalreport}, with grammatically correct nonsense text concatenated to match the document length distribution of FineWeb2. \emph{T} denotes the threshold used to retain the top 10\% of data in our experiments. The colored area represents the proportion of retained documents.}
\end{figure}

\textbf{Sample 1: Structured Educational Essay (Poetry)}
\begin{quote}
    \small
    \textbf{Text:} la poésie- réalité/ poésie: forme d'évasion du réel :Introduction La poésie est un genre littéraire qui permet d'exprimer des sentiments... : Thèse La poésie peut être considérée comme une forme d'évasion du réel... : Antithèse Cependant, la poésie peut également être considérée comme une représentation de la réalité... :Conclusion En fin de compte, la poésie peut être considérée à la fois comme une forme d'évasion du réel et une représentation de la réalité... \\
    \textbf{Translation:} Poetry-reality/poetry: a form of escape from reality :Introduction Poetry is a literary genre that allows the expression of feelings... :Thesis Poetry can be considered a form of escape from reality because it allows the author and reader to escape... :Antithesis However, poetry can also be considered a representation of reality... :Conclusion Ultimately, poetry can be considered both a form of escape from reality and a representation of reality... \\
    \textbf{HQ Rank:} 18,773,069 (Score: 0.1301) $\rightarrow$ \textbf{ML Rank:} 10,298 (Score: 0.9988) \\
    \textbf{Shift:} +18,762,771
\end{quote}
\textbf{Analysis:} The monolingual \textit{HQ} classifier failed to prioritize this highly structured educational essay. In contrast, the \textit{ML} classifier correctly identified it as high-quality content. This suggests that Multilingual training sensitizes the model to universal academic markers (like "Introduction," "Thesis," and "Conclusion") which appear across many languages in the MKC+ pool.
\newline
\newline
\textbf{Sample 2: Grammatically Fluent Nonsense}
\begin{quote}
    \small
    \textbf{Text:} Cela façon dont la personne a une bouteille de vie dans la . Il s'agit d'attraction simplement pas sembler le problème avec vous mènera probablement vous aimez pas être exactement . Et pleine forme auprès de ce que c'est d'abord, les cadres supérieurs et plus faibles qui ne fonctionne de compliments semblerait être. \\
    \textbf{Translation:} This way in which the person has a bottle of life in the . It is a matter of attraction simply not to seem the problem with you will probably lead you do not like being exactly . And in great shape with what it is first, the senior and weaker executives who does not work of compliments would seem to be. \\
    \textbf{HQ Rank:} 65,072 (Score: 0.9959) $\rightarrow$ \textbf{ML Rank:} 18,768,564 (Score: 0.0199) \\
    \textbf{Shift:} -18,703,492
\end{quote}
\textbf{Analysis:} This text uses correct French words and localized syntax, but the meaning is total nonsense (e.g., ``a bottle of life in the''). The monolingual \textit{HQ} model was fooled by the surface-level fluency, but the \textit{ML} model correctly identified it as noise. This indicates that multilingual embeddings help the model verify semantic coherence, as ``nonsense'' rarely aligns well across different languages in latent space. In Figure \ref{fig:retention_of_nonsense_text} we see that this behaviour generalizes to Spanish in much less pronounced way, but does not hold for Chinese. 

\subsection{Romance (No-French) Classifier}

\textbf{Sample 3: Formal/Liturgical Text (Psalms)}
\begin{quote}
    \small
    \textbf{Text:} |1||2||3|... 1De David. Éternel, je me tourne vers toi, 2mon Dieu, en toi je me confie. Que je ne sois pas couvert de honte! Que mes ennemis ne se réjouissent pas à mon sujet!... 4Éternel, fais-moi connaître tes voies, enseigne-moi tes sentiers! 5Conduis-moi dans ta vérité... \\
    \textbf{Translation:} |1||2||3|... 1Of David. Eternal, I turn toward you, 2my God, in you I trust. Let me not be covered in shame! Let my enemies not rejoice over me!... 4Eternal, make me know your ways, teach me your paths! 5Lead me in your truth... \\
    \textbf{Baseline Rank:} 332,619,231 (Score: 0.0000) $\rightarrow$ \textbf{No-Fra Rank:} 82,936 (Score: 0.9942) \\
    \textbf{Shift:} +332,536,295
\end{quote}
\textbf{Analysis:} A classifier trained on other Romance languages (Spanish, Italian, etc.) but \textit{not} French significantly prioritized this liturgical text. This proves that formal registers and religious signatures are highly conserved across language families. The ``quality'' of this register is recognized zero-shot across languages.
\newline
\newline
\textbf{Sample 4: Transient News (Sports)}
\begin{quote}
    \small
    \textbf{Text:} Myriam Soumaré s’est qualifiée pour les demi-finales du 200m des Championnats du monde d’athlétisme, ce jeudi à Moscou, en terminant troisième de sa série en 22’’83. La sprinteuse française, championne d’Europe de la discipline en 2010, tentera de se qualifier pour la finale... \\
    \textbf{Translation:} Myriam Soumaré qualified for the semi-finals of the 200m at the World Athletics Championships this Thursday in Moscow, finishing third in her heat in 22''83. The French sprinter, European champion in the discipline in 2010, will attempt to qualify for the final... \\
    \textbf{Baseline Rank:} 13,957,851 (Score: 0.3891) $\rightarrow$ \textbf{No-Fra Rank:} 318,322,669 (Score: 0.0000) \\
    \textbf{Shift:} -304,364,818
\end{quote}
\textbf{Analysis:} While informative, this snippet was heavily penalized by the Romance-transfer model. This confirms our hypothesis that cross-family transfer pushes the model toward ``Encyclopedic'' quality anchors (like Wikipedia) and away from the more ``common'' reporting found in general web crawls.

\subsection{Nordic Classifier}

\textbf{Sample 5: Public Park Information}
\begin{quote}
    \small
    \textbf{Text:} Square de la place de la Réunion. Horaires: Ouvert en ce moment |jeudi 21/03||08:00 à 18:00|... En 1849, le village dit « Grand Charonne » rejoint le hameau le « Petit Charonne »... Ce jardin contemporain a été éco labélisé en 2012... Le hêtre pourpre au feuillage rouge brun crée une continuité avec les coloris pourpres du fossé humide. \\
    \textbf{Translation:} Square of the Place de la Réunion. Hours: Open now |Thursday 03/21||08:00 to 18:00|... In 1849, the village known as "Grand Charonne" joined the hamlet of "Petit Charonne"... This contemporary garden was eco-labeled in 2012... The purple beech with red-brown foliage creates continuity with the purple colors of the wet ditch. \\
    \textbf{HQ Rank:} 327,597,576 (Score: 0.0001) $\rightarrow$ \textbf{Nordic Rank:} 8,268,964 (Score: 0.3525) \\
    \textbf{Shift:} +319,328,612
\end{quote}
\textbf{Analysis:} This document is heavily ``noisy,'' starting with long tables of opening hours and ending with Twitter handles. The monolingual \textit{HQ} baseline likely penalized the initial boilerplate so severely that it discarded the entire document. However, the \textit{Nordic} classifier identifies the knowledge-dense middle section (historical facts and detailed botanical descriptions). This suggests that cross-family transfer models can be more robust to localized boilerplate, focusing instead on the global density of information.
\newline
\newline
\textbf{Sample 6: E-commerce Boilerplate (Empty Cart)}
\begin{quote}
    \small
    \textbf{Text:} Toutes les catégories Votre panier est vide! Index des marques: 0 - 9 A B C D E... Ce produit est en rupture de stock. Vous pouvez remplir ce formulaire pour être notifié... \\
    \textbf{Translation:} All categories Your cart is empty! Brand index: 0-9 A B C D E... This product is out of stock. You can fill out this form to be notified... \\
    \textbf{Baseline Rank:} 6,322,191 (Score: 0.6707) $\rightarrow$ \textbf{Nordic Rank:} 320,914,784 (Score: 0.0000) \\
    \textbf{Shift:} -314,592,593
\end{quote}
\textbf{Analysis:} Even without knowing French, the \textit{Nordic} model identifies the structural signature of low-utility e-commerce pages (e.g., brand lists, empty cart messages). These patterns are cross-lingual, allowing the model to effectively filter web noise in languages it has never seen, compared to the HQ baseline, which gave it a high score due to its fluent writing.

\section{Analysis of Q3 Hard Negatives}
\label{appendix:samplesq3}

The Q3 sampling strategy selects documents that score in the 50th-75th percentile of the baseline distribution. These samples are critical for refining the decision boundary of the classifier.

\textbf{Q3 negative sample: Administrative School Snippets}
\begin{quote}
    \small
    \textbf{Text:} Publié dans Vie des écoles le 16.09.12. Les élèves des écoles Léonard De Vinci collectent les journaux (quotidiens nationaux et régionaux). Publié dans Vie des écoles le 18.09.10. Consulter les différentes commissions et comités de l'année scolaire 2012-2013. La grande affaire de la rentrée 2008/2009 à l'école élémentaire Léonard de Vinci a été la mise en place de l'aide personnalisée pour venir en aide aux enfants en difficultés. Cette année, les effectifs sont en hausse : 118 élèves sont inscrits à l'école maternelle. \\
    \textbf{Translation:} Published in School Life on 16.09.12. Students from the Léonard De Vinci schools collect newspapers (national and regional dailies). Published in School Life on 18.09.10. Consult the various commissions and committees for the 2012-2013 school year. The main development of the 2008/2009 school year at the Léonard de Vinci elementary school was the implementation of personalized support to assist children with difficulties. This year, enrollment is rising: 118 students are enrolled in the nursery school. \\
    \textbf{Label:} Negative (Q3 Sample)
\end{quote}
\textbf{Analysis:} This document is a perfect example of a useful negative. It is flawlessly written, uses proper punctuation, and contains no ``web noise'' like ads or code. However, its informational content is hyper-local, administrative, and transient (referring to school enrollment numbers and newspaper drives from 2008–2012). While a standard classifier might be tempted to rank this highly because it contains educational keywords like \textit{école} (school) and \textit{Léonard De Vinci}, the Q3 strategy teaches the model that \textbf{fluency does not equal informational depth}. By using such samples as negatives, we force the classifier to look past surface-level grammar and prioritize documents with actual semantic or scientific weight.

\section{Megatron Config}
\label{appendix:meg_config}
To train the 1B models, we use the Apertus tokenizer, the Megatron LM library~\citep{shoeybi2020megatronlmtrainingmultibillionparameter}, a batch size of 2.06M tokens, a learning rate of 0.00015, the AdEMAMix optimizer~\citep{pagliardini2024ademamixoptimizerbetterfaster}, 50,000 training steps with 2000 warmup steps, and a WSD learning rate schedule ~\citep{hägele2024scalinglawscomputeoptimaltraining}. 
These models were trained using 84 NVIDIA GH200 chips. The full configuration for these models is displayed in Table~\ref{tab:training_config}). This training pipeline is very similar to what has been done in~\citep{fineweb2HQ}.
\begin{longtable}{ll}
\caption{Training Configuration for Apertus 1B Model}
\label{tab:training_config} \\
\multicolumn{1}{c}{\bf Parameter} &
\multicolumn{1}{c}{\bf Value}\\
\midrule
\endfirsthead
\multicolumn{2}{c}%
{\tablename\ \thetable\ -- \textit{Continued from previous page}} \\ \\
\multicolumn{1}{c}{\bf Parameter} &
\multicolumn{1}{c}{\bf Value}\\
\hline \\
\endhead
\hline
\multicolumn{2}{r}{\textit{Continued on next page}} \\
\endfoot
\hline
\endlastfoot
\multicolumn{2}{l}{\textit{Model Architecture}} \\
Number of Layers & 16 \\
Hidden Size & 2048 \\
FFN Hidden Size & 12288 \\
Number of Attention Heads & 32 \\
Number of Query Groups & 8 \\
Maximum Position Embeddings & 4096 \\
Position Embedding Type & RoPE \\
RoPE Base & 500000 \\
RoPE Scaling Factor & 32 \\
Normalization & RMSNorm \\
Activation Function & XieLU \\
\hline
\multicolumn{2}{l}{\textit{Training Configuration}} \\
Micro Batch Size & 3 \\
Global Batch Size & 504 \\
Sequence Length & 4096 \\
Total Training Steps & 50,000 \\
Total Tokens & $\sim$103B \\
Checkpoint Interval & 2,000 steps \\
\hline
\multicolumn{2}{l}{\textit{Optimization}} \\
Optimizer & AdEMAMix \\
Learning Rate & 0.00015 \\
Minimum Learning Rate & 0.0 \\
LR Schedule & WSD (1-sqrt decay) \\
Warmup Steps & 2,000 \\
WSD Decay Steps & 10,000 \\
Weight Decay & 0.1 \\
Gradient Clipping & 0.1 \\
Adam $\beta_1$ & 0.9 \\
Adam $\beta_2$ & 0.999 \\
AdEMAMix $\alpha$ & 8 \\
AdEMAMix $\beta_3$ & 0.9999 \\
AdEMAMix $\beta_3$ Warmup & 100,000 \\
AdEMAMix $\alpha$ Warmup & 100,000 \\
\hline
\multicolumn{2}{l}{\textit{Regularization}} \\
Attention Dropout & 0.0 \\
Hidden Dropout & 0.0 \\
\hline
\multicolumn{2}{l}{\textit{Infrastructure}} \\
Number of Nodes & 21 \\
GPUs per Node & 4 \\
Total GPUs & 84 \\
Tensor Parallelism & 1 \\
Pipeline Parallelism & 1 \\
Precision & BF16 \\
\hline
\multicolumn{2}{l}{\textit{Additional Features}} \\
Tokenizer & swiss-ai/Apertus-70B-2509 \\
Goldfish Loss ($k$, $h$) & 50, 50 \\
Cross-document Attention & Enabled \\
QK LayerNorm & Enabled \\
Seed & 28 \\
\end{longtable}

\section{Ranking Procedure for approaches}
For each language, we evaluate filtering strategies by training 1B parameter Apertus models on their filtered outputs and benchmarking on language-appropriate tasks (detailed in Appendix~\ref{appendix:benchmarks}). We rank methods by their performance on each individual benchmark, then compute the average rank across all benchmarks for that language. A lower average rank indicates better overall performance. This ranking approach is robust to scale differences across benchmarks and emphasizes consistency across diverse evaluation tasks.
We also include mean normalized accuracy as a metric, as it gives us more quantitative insight for the performance gain of each method. 

\section{Benchmarks by Language}
\label{appendix:benchmarks}

\subsection{French}
The following benchmarks were used to evaluate model performance on French:
\begin{itemize}
    \item ARC-Challenge~\citep{ARC}
    \item Belebele~\citep{beleble}
    \item Global-MMLU~\citep{singh2025globalmmluunderstandingaddressing}
    \item HellaSwag~\citep{zellers2019hellaswag, dac2023okapi}
    \item Include-Base-44~\citep{include}
    \item Multilingual MMLU~\citep{alexandra_institute_2025}
    \item XNLI~\citep{xnli}
    \item XWinograd~\citep{muennighoff2022crosslingual, tikhonov2021heads}
\end{itemize}

\subsection{Arabic}
The following benchmarks were used to evaluate model performance on Arabic:
\begin{itemize}
    \item ARC-Easy~\citep{ARC}
    \item AlGhafa PIQA-MT~\citep{almazrouei-etal-2023-alghafa}
    \item AlGhafa RACE~\citep{alghafa2023}
    \item AlGhafa SciQ~\citep{alghafa2023}
    \item ARC-Challenge~\citep{ARC}
    \item Belebele~\citep{beleble}
    \item Global-MMLU~\citep{singh2025globalmmluunderstandingaddressing}
    \item HellaSwag~\citep{zellers2019hellaswag, dac2023okapi}
    \item Include-Base-44~\citep{include}
    \item Multilingual MMLU~\citep{alexandra_institute_2025}
    \item AlGhafa PIQA~\citep{bisk2019piqareasoningphysicalcommonsense, alghafa2023}
    \item XNLI~\citep{xnli}
    \item XStoryCloze~\citep{DBLP:journals/corr/abs-2112-10668}
\end{itemize}

\subsection{Spanish}
The following benchmarks were used to evaluate model performance on Spanish:
\begin{itemize}
    \item ARC-Challenge~\citep{ARC}
    \item Belebele~\citep{beleble}
    \item Global-MMLU~\citep{singh2025globalmmluunderstandingaddressing}
    \item HellaSwag~\citep{zellers2019hellaswag, dac2023okapi}
    \item Include-Base-44~\citep{include}
    \item Multilingual MMLU~\citep{alexandra_institute_2025}
    \item XNLI~\citep{xnli}
\end{itemize}

\subsection{Chinese}
The following benchmarks were used to evaluate model performance on Chinese:
\begin{itemize}
    \item Agieval Cn~\citep{zhong2023agieval, ling-etal-2017-program, hendrycksmath2021, Liu2020LogiQAAC, zhong2019jec, Wang2021FromLT}
    \item ARC~\citep{ARC}
    \item Belebele~\citep{beleble}
    \item Ceval-valid~\citep{huang2023ceval}
    \item Cmmlu~\citep{li2024cmmlumeasuringmassivemultitask}
    \item Global-MMLU~\citep{singh2025globalmmluunderstandingaddressing}
    \item Include-Base-44~\citep{include}
    \item Multilingual MMLU~\citep{alexandra_institute_2025}
    \item PAWS-X~\citep{yang2019pawsxcrosslingualadversarialdataset}
    \item Xcopa~\citep{ponti2020xcopa, roemmele2011choice}
    \item XNLI~\citep{xnli}
    \item XStoryCloze~\citep{DBLP:journals/corr/abs-2112-10668}
    \item XWinograd~\citep{muennighoff2022crosslingual, tikhonov2021heads}
\end{itemize}

\section{The Role of Scale and Seed Variance}
\label{appendix:seed}
We have investigated the addition of languages and the curation of negative samples. However, we have to ask ourselves about their stability. 
If we were to supply different positive/negative samples, we would expect to get very similar results. To test this hypothesis, we vary the sampling seed of the classifier. This would result in selecting different samples from our positive anchors, but also from FineWeb2. This experiment is labeled as \textbf{HQ seed}.
\begin{table}[t]
\caption{Comparison of the HQ baseline against a version trained with a different sampling seed (HQ seed) on Arabic. The variation in results across tasks suggests that in lower-resource settings, the specific documents selected for the anchor set can significantly impact the classifier's decision boundary.}
\label{tab:seed_arabic}
\begin{center}
\begin{tabular}{lccc}
\multicolumn{1}{c}{\bf Benchmark} &
\multicolumn{1}{c}{\bf No filtering} &
\multicolumn{1}{c}{\bf HQ} &
\multicolumn{1}{c}{\bf HQ seed}\\
\midrule
ARC-Easy & 0.2716 & \textbf{0.2898} & 0.2838 \\
AlGhafa PIQA-MT & \textbf{0.5194} & 0.5145 & 0.5095 \\
AlGhafa RACE & 0.2715 & 0.2775 & \textbf{0.2879} \\
AlGhafa SciQ & 0.4261 & 0.4432 & \textbf{0.4492} \\
ARC-Challenge & 0.2678 & 0.2660 & \textbf{0.2720} \\
Belebele\_c & 0.3222 & 0.3233 & \textbf{0.3289} \\
GMMLU\_c & 0.2475 & \textbf{0.2575} & 0.2450 \\
HellaSwag & \textbf{0.3909} & 0.3873 & 0.3909 \\
Include\_c & \textbf{0.3025} & 0.2681 & 0.2844 \\
M\_MMLU\_c & \textbf{0.2669} & 0.2614 & 0.2653 \\
AlGhafa PIQA & \textbf{0.6132} & 0.6088 & 0.6126 \\
XNLI & \textbf{0.3349} & 0.3309 & 0.3317 \\
XStoryCloze & 0.5903 & 0.5923 & \textbf{0.5930} \\
\hline
\textbf{Aggregate acc\_norm} & 0.3711 & 0.3708 & \textbf{0.3734} \\
\textbf{Average rank} & 1.92 & 2.31 & \textbf{1.69} \\
\hline
\end{tabular}
\end{center}
\end{table}

\begin{table}[t]
\caption{Comparison of standard HQ against variants with different classifier seeds, LLM training seeds, and larger positive anchor sets (HQ all) on French. Results indicate that variance in data selection (classifier seed) has a larger impact on downstream performance than the stochasticity of the LLM training itself.}
\label{benchmark:french_seed}
\begin{center}
\begin{tabular}{lccccc}
\multicolumn{1}{c}{\bf Benchmark} &
\multicolumn{1}{c}{\bf No filtering} &
\multicolumn{1}{c}{\bf HQ} &
\multicolumn{1}{c}{\bf HQ (seed)} &
\multicolumn{1}{c}{\bf \makecell{HQ seed \\ LLM Training}} &
\multicolumn{1}{c}{\bf HQ (all)}\\
\midrule
ARC-Challenge & 0.2891 & 0.3071 & \textbf{0.3216} & 0.2985 & 0.3080 \\
Belebele\_c & 0.3444 & 0.3511 & 0.3500 & 0.3389 & \textbf{0.3544} \\
GMMLU\_c & 0.2625 & 0.2925 & 0.2900 & \textbf{0.3075} & 0.2875 \\
HellaSwag & \textbf{0.4883} & 0.4748 & 0.4761 & 0.4774 & 0.4773 \\
Include\_c & 0.3866 & \textbf{0.4153} & 0.4057 & 0.4129 & 0.4105 \\
M\_MMLU\_c & 0.2831 & 0.2945 & 0.2944 & 0.2946 & \textbf{0.2949} \\
XNLI & 0.4707 & 0.4855 & 0.4823 & \textbf{0.4904} & 0.4695 \\
XWinograd & 0.6386 & \textbf{0.6506} & 0.5904 & 0.6265 & 0.6386 \\
\hline
\textbf{Aggregate acc\_norm} & 0.3954 & \textbf{0.4089} & 0.4013 & 0.4058 & 0.4051 \\
\textbf{Average rank} & 3.88 & \textbf{2.38} & 3.38 & 2.62 & 2.62 \\
\hline
\end{tabular}
\end{center}
\end{table}
Tables~\ref{tab:seed_arabic} and~\ref{benchmark:french_seed} analyze the stability of quality filtering with respect to sampling variance and training scale. 
While multilingual pooling improves overall rank stability, our ablations show that downstream LLM performance remains sensitive to the classifier's initial sampling seed. Changing the seed shifts aggregate accuracy by ~0.3\% in Arabic and ~0.8\% in French (Tables \ref{tab:seed_arabic} and \ref{benchmark:french_seed}). This variance suggests that the specific subset of examples used to define the quality boundary heavily influences which knowledge domains are selected.

To mitigate this variance, we examine two complementary strategies: increasing positive sample coverage by using all available high-quality anchors (HQ all), and varying the LLM training seed independently from the filtering process. While altering the LLM seed introduces minor variability, the dominant source of instability arises from the classifier sampling process itself.

Using the full positive set moderately improves robustness but does not consistently outperform standard HQ filtering, suggesting diminishing returns from scale alone. 

This result is counterintuitive if we assume that ``more data is better.'' We hypothesize that this degradation is due to an informational saturation effect: by providing the classifier with the entire, unfiltered anchor pool, we likely introduced a higher ratio of ``non-helpful'' or marginal samples that are present in the positive datasets but lack strong educational signal. This prevents the classifier from establishing a sharp decision boundary between truly high-quality content and baseline web text. This suggests that representative, curated sampling is a more effective strategy for training quality filters than just increasing the number of training samples.
These results motivate multilingual and bootstrapped approaches introduced in the paper, which effectively average over topic and language variability to yield more stable quality signals.

\section{Addition of MKC-e datasets.}
To make our classifier highly multilingual beyond the coverage given by the Aya Collection~\citep{ayadataset}, we extend the MKC+ pool with additional datasets (resulting in the MKC-e data). A hypothesis is that, beyond allowing us to train on more languages, this addition provides more topics and diversity, which can improve the performance of our filtering. In order to test this, we train some monolingual classifiers on the MKC-e data and evaluate them using the training of the 1B Apertus. Results are displayed in Tables~\ref{benchmark:spanishmkce}, \ref{benchmark:frenchmkce}, and \ref{benchmark:arabicmkce}. While the training on MKC-e improves results for Spanish by almost 0.5\% in terms of normalized accuracy, we find that the results are more nuanced for French, where the improvement is of 0.1\%. However, the Arabic MKC-e classifier seems to 
perform worse than the ``No filtering'' baseline, which suggests adding this new data pool provides mixed results depending on the targeted language.

\begin{table}[t]

\caption{Evaluation of the monolingual HQ baseline versus a classifier trained on the expanded MKC-e pool on Spanish. The inclusion of instruction and synthetic data improves aggregate normalized accuracy by nearly 0.5\%, indicating high utility for Spanish domain coverage.}
\label{benchmark:spanishmkce}

\begin{center}
\begin{tabular}{lccc}
\multicolumn{1}{c}{\bf Benchmark} &
\multicolumn{1}{c}{\bf No filtering} &
\multicolumn{1}{c}{\bf HQ} &
\multicolumn{1}{c}{\bf MKC-e}\\
\midrule
ARC-Challenge & 0.2991 & 0.3077 & \textbf{0.3291} \\
Belebele\_c   & 0.3422 & \textbf{0.3533} & 0.3378 \\
GMMLU\_c      & 0.3100 & 0.3150 & \textbf{0.3200} \\
HellaSwag     & 0.5006 & \textbf{0.5263} & 0.5075 \\
Include\_c    & 0.3491 & 0.3727 & \textbf{0.3964} \\
M\_MMLU\_c    & 0.2814 & 0.2985 & \textbf{0.3137} \\
XNLI          & \textbf{0.4618} & 0.4538 & 0.4566 \\
\hline
\textbf{Aggregate acc\_norm} & 0.3635 & 0.3753 & \textbf{0.3801} \\
\textbf{Average rank} & 2.57 & 1.86 & \textbf{1.57} \\
\hline
\end{tabular}
\end{center}

\end{table}

\begin{table}[t]

\caption{Comparison of standard HQ versus the MKC-e anchor set. The extended data provides a marginal gain in aggregate accuracy (+0.1\%) but significantly improves performance on specific benchmarks like ARC-Challenge and XWinograd.}
\label{benchmark:frenchmkce}

\begin{center}
\begin{tabular}{lccc}
\multicolumn{1}{c}{\bf Benchmark} &
\multicolumn{1}{c}{\bf No filtering} &
\multicolumn{1}{c}{\bf HQ} &
\multicolumn{1}{c}{\bf MKC-e}\\
\midrule
ARC-Challenge & 0.2891 & 0.3071 & \textbf{0.3259} \\
Belebele\_c   & 0.3444 & 0.3511 & \textbf{0.3533} \\
GMMLU\_c      & 0.2625 & \textbf{0.2925} & 0.2650 \\
HellaSwag     & \textbf{0.4883} & 0.4748 & 0.4647 \\
Include\_c    & 0.3866 & 0.4153 & \textbf{0.4296} \\
M\_MMLU\_c    & 0.2831 & \textbf{0.2945} & 0.2929 \\
XNLI          & 0.4707 & \textbf{0.4855} & 0.4735 \\
Xwinograd     & 0.6386 & 0.6506 & \textbf{0.6747} \\
\hline
\textbf{Aggregate acc\_norm} & 0.3954 & 0.4089 & \textbf{0.4099} \\
\textbf{Average rank} & 2.75 & \textbf{1.62} & \textbf{1.62} \\
\hline
\end{tabular}
\end{center}

\end{table}

\begin{table}[t]

\caption{Arabic performance with extended anchors (MKC-e). Unlike the Romance languages, adding the MKC-e pool to Arabic filtering slightly degrades aggregate performance. This suggests that the instruction-tuning signal in MKC-e may not align as cleanly with "educational quality" for  Arabic.}
\label{benchmark:arabicmkce}

\begin{center}
\begin{tabular}{lccc}
\multicolumn{1}{c}{\bf Benchmark} &
\multicolumn{1}{c}{\bf No filtering} &
\multicolumn{1}{c}{\bf HQ} &
\multicolumn{1}{c}{\bf MKC-e}\\
\midrule
ARC-Easy        & 0.2716 & \textbf{0.2898} & 0.2728 \\
AlGhafa PIQA-MT & \textbf{0.5194} & 0.5145 & 0.4970 \\
AlGhafa RACE    & 0.2715 & \textbf{0.2775} & 0.2753 \\
AlGhafa SciQ    & 0.4261 & \textbf{0.4432} & 0.4422 \\
ARC-Challenge   & \textbf{0.2678} & 0.2660 & 0.2643 \\
Belebele\_c     & 0.3222 & 0.3233 & \textbf{0.3244} \\
GMMLU\_c        & 0.2475 & \textbf{0.2575} & 0.2550 \\
HellaSwag       & \textbf{0.3909} & 0.3873 & 0.3882 \\
Include\_c      & \textbf{0.3025} & 0.2681 & 0.2953 \\
M\_MMLU\_c      & \textbf{0.2669} & 0.2614 & 0.2628 \\
AlGhafa PIQA    & \textbf{0.6132} & 0.6088 & 0.6121 \\
XNLI            & \textbf{0.3349} & 0.3309 & 0.3349 \\
Xstorycloze     & 0.5903 & \textbf{0.5923} & 0.5890 \\
\hline
\textbf{Aggregate acc\_norm} & \textbf{0.3711} & 0.3708 & 0.3703 \\
\textbf{Average rank} & \textbf{1.85} & 2.00 & 2.08 \\
\hline
\end{tabular}
\end{center}

\end{table}

\begin{table}[t]
\caption{Spanish comprehensive benchmark results. Final comparison of all filtering strategies. ML (Q3) achieves the best average rank, showing that the combination of multilingual signal and sharpened decision boundaries is optimal for Spanish.}
\label{benchmark:spanish}
\begin{center}
\begin{tabular}{lccccccc}
\multicolumn{1}{c}{\bf Benchmark} &
\multicolumn{1}{c}{\bf No filtering} &
\multicolumn{1}{c}{\bf HQ} &
\multicolumn{1}{c}{\bf HQ seed} &
\multicolumn{1}{c}{\bf MKC-e} &
\multicolumn{1}{c}{\bf ML} &
\multicolumn{1}{c}{\bf \makecell{ML\\(15\%)}} &
\multicolumn{1}{c}{\bf \makecell{ML\\(Q3)}}\\
\midrule
ARC-Challenge & 0.2991 & 0.3077 & 0.3274 & 0.3291 & 0.3248 & \textbf{0.3436} & 0.3282 \\
Belebele\_c & 0.3422 & \textbf{0.3533} & 0.3422 & 0.3378 & 0.3456 & 0.3500 & 0.3533 \\
GMMLU\_c & 0.3100 & 0.3150 & \textbf{0.3375} & 0.3200 & 0.3250 & 0.3225 & 0.3275 \\
HellaSwag & 0.5006 & 0.5263 & 0.5219 & 0.5075 & 0.5310 & \textbf{0.5440} & 0.5372 \\
Include\_c & 0.3491 & 0.3727 & 0.3782 & 0.3964 & 0.3891 & \textbf{0.4036} & 0.4000 \\
M\_MMLU\_c & 0.2814 & 0.2985 & 0.3073 & \textbf{0.3137} & 0.3050 & 0.3080 & 0.3083 \\
XNLI & 0.4618 & 0.4538 & 0.4707 & 0.4566 & \textbf{0.4783} & 0.4562 & 0.4667 \\
\hline
\textbf{Aggregate acc\_norm} & 0.3635 & 0.3753 & 0.3836 & 0.3801 & 0.3855 & \textbf{0.3897} & 0.3887 \\
\textbf{Average rank} & 6.29 & 5.14 & 3.71 & 4.14 & 3.57 & 2.71 & \textbf{2.14} \\
\hline
\end{tabular}
\end{center}
\end{table}

\section{Global Benchmark results}
\clearpage
\begin{table}[t]
\caption{Chinese comprehensive benchmark results. Final comparison of all filtering strategies. The ML strategy secures the best rank and aggregate accuracy, demonstrating that multilingual signal provides the most robust quality filter for Chinese logographic text.}
\label{benchmark:chinese}
\begin{center}
\begin{tabular}{lccccccc}
\multicolumn{1}{c}{\bf Benchmark} &
\multicolumn{1}{c}{\bf No filtering} &
\multicolumn{1}{c}{\bf HQ} &
\multicolumn{1}{c}{\bf HQ seed} &
\multicolumn{1}{c}{\bf MKC-e} &
\multicolumn{1}{c}{\bf ML} &
\multicolumn{1}{c}{\bf \makecell{ML\\(15\%)}} &
\multicolumn{1}{c}{\bf \makecell{ML\\(Q3)}}\\
\midrule
Agieval Cn & 0.3618 & 0.3644 & 0.3625 & 0.3609 & 0.3457 & 0.3497 & \textbf{0.3658} \\
ARC & 0.2855 & 0.3145 & 0.3000 & 0.3162 & \textbf{0.3171} & 0.3068 & 0.3085 \\
Belebele\_c & 0.3011 & 0.3200 & 0.3256 & 0.3433 & 0.3222 & 0.3356 & \textbf{0.3478} \\
Ceval-valid & 0.2288 & 0.2489 & 0.2444 & 0.2556 & \textbf{0.2615} & 0.2370 & 0.2467 \\
Cmmlu\_c & 0.3206 & 0.3471 & 0.3445 & 0.3581 & 0.3608 & 0.3466 & \textbf{0.3688} \\
GMMLU\_c & 0.2800 & 0.3075 & 0.3175 & \textbf{0.3200} & 0.3200 & 0.3025 & 0.3200 \\
Include\_c & 0.3468 & 0.3523 & 0.3523 & 0.3541 & 0.3541 & \textbf{0.3780} & 0.3670 \\
M MMLU\_c & 0.2772 & 0.2940 & 0.2937 & 0.2959 & \textbf{0.2987} & 0.2927 & 0.2978 \\
PAWS & 0.5520 & 0.5535 & 0.5360 & 0.5435 & \textbf{0.5610} & 0.5570 & 0.5480 \\
Xcopa & 0.5860 & 0.5920 & \textbf{0.6200} & 0.5940 & 0.6080 & 0.6160 & 0.6020 \\
XNLI & 0.3546 & 0.4072 & 0.3779 & 0.4056 & \textbf{0.4189} & 0.3980 & 0.3627 \\
Xstorycloze & 0.6559 & 0.6625 & 0.6605 & 0.6618 & 0.6625 & 0.6678 & \textbf{0.6750} \\
XWinograd & 0.6806 & 0.6825 & \textbf{0.6964} & 0.6925 & 0.6766 & 0.6865 & 0.6667 \\
\hline
\textbf{Aggregate acc\_norm} & 0.4024 & 0.4190 & 0.4178 & 0.4232 & \textbf{0.4236} & 0.4211 & 0.4213 \\
\textbf{Average rank} & 6.38 & 3.85 & 4.46 & 3.23 & \textbf{2.69} & 3.92 & 3.00 \\
\hline
\end{tabular}
\end{center}
\end{table}

\begin{table}[t]
\caption{Arabic comprehensive benchmark results. Final comparison of all filtering strategies. The ML strategy (56\% retention) remains the dominant approach, while aggressive filtering (10-20\%) consistently underperforms regardless of the classifier used.}
\label{benchmark:arabic}
\begin{center}
\resizebox{\textwidth}{!}{
\begin{tabular}{lcccccccccc}
\multicolumn{1}{c}{\bf Benchmark} &
\multicolumn{1}{c}{\bf \makecell{No\\Filtering}} &
\multicolumn{1}{c}{\bf HQ} &
\multicolumn{1}{c}{\bf \makecell{HQ\\ seed}} &
\multicolumn{1}{c}{\bf MKC-e} &
\multicolumn{1}{c}{\bf ML} &
\multicolumn{1}{c}{\bf \makecell{ML\\(Q3)}} &
\multicolumn{1}{c}{\bf \makecell{HQ\\(10\%)}} &
\multicolumn{1}{c}{\bf \makecell{HQ\\(20\%)}} &
\multicolumn{1}{c}{\bf \makecell{ML\\(10\%)}} &
\multicolumn{1}{c}{\bf \makecell{ML\\(20\%)}}\\
\midrule
ARC-Easy & 0.2716 & \textbf{0.2898} & 0.2838 & 0.2728 & 0.2855 & 0.2750 & 0.2720 & 0.2771 & 0.2741 & 0.2762 \\
AlGhafa PIQA-MT & \textbf{0.5194} & 0.5145 & 0.5095 & 0.4970 & 0.5112 & 0.5085 & 0.5019 & 0.5090 & 0.4948 & 0.5128 \\
AlGhafa RACE & 0.2715 & 0.2775 & 0.2879 & 0.2753 & \textbf{0.2883} & 0.2792 & 0.2747 & 0.2834 & 0.2765 & 0.2786 \\
AlGhafa SciQ & 0.4261 & 0.4432 & 0.4492 & 0.4422 & 0.4503 & 0.4322 & 0.4503 & 0.4171 & \textbf{0.4573} & 0.4563 \\
ARC-Challenge & 0.2678 & 0.2660 & 0.2720 & 0.2643 & \textbf{0.2797} & 0.2601 & 0.2669 & 0.2772 & 0.2712 & 0.2626 \\
Belebele\_c & 0.3222 & 0.3233 & 0.3289 & 0.3244 & 0.3122 & \textbf{0.3444} & 0.3378 & 0.2944 & 0.3278 & 0.3256 \\
GMMLU\_c & 0.2475 & 0.2575 & 0.2450 & 0.2550 & 0.2700 & 0.2575 & 0.2475 & 0.2550 & \textbf{0.2725} & 0.2525 \\
HellaSwag & 0.3909 & 0.3873 & 0.3909 & 0.3882 & \textbf{0.3925} & 0.3870 & 0.3592 & 0.3728 & 0.3747 & 0.3857 \\
Include\_c & 0.3025 & 0.2681 & 0.2844 & 0.2953 & \textbf{0.3043} & 0.2754 & 0.2717 & 0.2772 & 0.2844 & 0.2790 \\
M\_MMLU\_c & 0.2669 & 0.2614 & 0.2653 & 0.2628 & 0.2646 & 0.2615 & 0.2652 & \textbf{0.2674} & 0.2618 & 0.2634 \\
AlGhafa PIQA & 0.6132 & 0.6088 & 0.6126 & 0.6121 & 0.6110 & 0.6072 & 0.5958 & 0.6023 & 0.6039 & \textbf{0.6143} \\
XNLI & 0.3349 & 0.3309 & 0.3317 & 0.3349 & 0.3349 & \textbf{0.3373} & 0.3325 & 0.3333 & 0.3325 & 0.3357 \\
XStoryCloze & 0.5903 & 0.5923 & \textbf{0.5930} & 0.5890 & 0.5811 & 0.5831 & 0.5877 & 0.5817 & 0.5784 & 0.5804 \\
\hline
\textbf{Aggregate acc\_norm} & 0.3711 & 0.3708 & 0.3734 & 0.3703 & \textbf{0.3758} & 0.3699 & 0.3664 & 0.3652 & 0.3700 & 0.3710 \\
\textbf{Average rank} & 5.00 & 5.77 & 4.08 & 5.85 & \textbf{3.46} & 5.85 & 6.92 & 6.08 & 6.08 & 5.15 \\
\hline
\end{tabular}
}
\end{center}
\end{table}

\begin{table}[t]
\caption{French comprehensive benchmark results. Detailed comparison of 12 distinct filtering strategies. While ML (15\%) achieves the highest absolute accuracy, the Q3 negatives and Nordic transfer models remain highly competitive, proving that quality can be captured through multiple distinct curation pathways.}
\label{benchmark:french}
\begin{center}
\resizebox{\textwidth}{!}{
\begin{tabular}{lcccccccccccc}
\multicolumn{1}{c}{\bf Benchmark} &
\multicolumn{1}{c}{\bf \makecell{No\\Filtering}} &
\multicolumn{1}{c}{\bf \makecell{HQ}} &
\multicolumn{1}{c}{\bf \makecell{HQ\\seed}} &
\multicolumn{1}{c}{\bf \makecell{HQ seed\\(LLM\\Training)}} &
\multicolumn{1}{c}{\bf \makecell{HQ\\all}} &
\multicolumn{1}{c}{\bf \makecell{Q3\\negatives}} &
\multicolumn{1}{c}{\bf \makecell{HQ\\Romance\\(No Fra)}} &
\multicolumn{1}{c}{\bf \makecell{MKC-e}} &
\multicolumn{1}{c}{\bf \makecell{Nordic}} &
\multicolumn{1}{c}{\bf \makecell{ML}} &
\multicolumn{1}{c}{\bf \makecell{ML\\(15\%)}} &
\multicolumn{1}{c}{\bf \makecell{ML\\(Q3)}}\\
\midrule
ARC-Challenge & 0.2891 & 0.3071 & 0.3216 & 0.2985 & 0.3080 & 0.3105 & 0.3054 & \textbf{0.3259} & 0.3157 & 0.3165 & 0.3139 & 0.3054 \\
Belebele\_c & 0.3444 & 0.3511 & 0.3500 & 0.3389 & 0.3544 & \textbf{0.3711} & 0.3689 & 0.3533 & 0.3422 & 0.3511 & 0.3611 & 0.3633 \\
GMMLU\_c & 0.2625 & 0.2925 & 0.2900 & \textbf{0.3075} & 0.2875 & 0.3025 & 0.2750 & 0.2650 & 0.3075 & 0.2900 & 0.2875 & 0.3075 \\
HellaSwag & 0.4883 & 0.4748 & 0.4761 & 0.4774 & 0.4773 & 0.4727 & 0.4908 & 0.4647 & 0.4673 & 0.4956 & \textbf{0.5135} & 0.4911 \\
Include\_c & 0.3866 & 0.4153 & 0.4057 & 0.4129 & 0.4105 & 0.4272 & 0.4272 & 0.4296 & 0.4535 & 0.4224 & \textbf{0.4726} & 0.4582 \\
M\_MMLU\_c & 0.2831 & 0.2945 & 0.2944 & 0.2946 & 0.2949 & 0.2992 & 0.2950 & 0.2929 & 0.2936 & 0.2927 & 0.2991 & \textbf{0.3007} \\
XNLI & 0.4707 & 0.4855 & 0.4823 & \textbf{0.4904} & 0.4695 & 0.4876 & 0.4827 & 0.4735 & 0.4783 & 0.4807 & 0.4807 & 0.4767 \\
XWinograd & 0.6386 & 0.6506 & 0.5904 & 0.6265 & 0.6386 & 0.6627 & 0.6024 & \textbf{0.6747} & 0.6627 & 0.6024 & 0.6386 & 0.6145 \\
\hline
\textbf{Aggregate acc\_norm} & 0.3954 & 0.4089 & 0.4013 & 0.4058 & 0.4051 & 0.4167 & 0.4059 & 0.4099 & 0.4151 & 0.4064 & \textbf{0.4209} & 0.4147 \\
\textbf{Average rank} & 9.88 & 6.38 & 7.62 & 6.75 & 7.38 & \textbf{4.00} & 6.00 & 6.88 & 6.12 & 6.50 & 4.12 & 4.62 \\
\hline
\end{tabular}
}
\end{center}
\end{table}
\section{Limitations}

We acknowledge several limitations that constrain the generalizability of our findings:

\paragraph{Statistical Rigor.}
Due to the computational cost of training 1B parameter models, all primary results are reported as single runs. Seed variance experiments (Tables~\ref{tab:seed_arabic} and~\ref{benchmark:french_seed}) show that changing the classifier sampling seed shifts aggregate accuracy by $\sim$0.3\% in Arabic and $\sim$0.8\% in French. For Arabic in particular, the ML gain over HQ ($\sim$0.5\%) is comparable to this noise range and should therefore be interpreted with caution. Future work should employ multiple seeds per condition to establish confidence intervals.

\paragraph{Embedding Space Interpretation.}
While our cross-lingual transfer results are empirically consistent, we cannot determine whether they reflect a genuinely abstract quality structure in the embedding space or shared formatting  across our positive anchor datasets, such as Wikipedia markup or instruction-tuning templates. Disentangling these mechanisms, for example through probing classifiers or controlled anchor set ablations, remains to be addressed in future work.
\paragraph{Limited Language Coverage.}
Our evaluation focuses on four languages, all of which have substantial representation in XLM-RoBERTa's pretraining corpus. Results may not generalize to extremely low-resource languages or underrepresented language families (e.g., Niger-Congo, Austronesian)

\paragraph{Embedding Model Dependence.}
All results rely on XLM-RoBERTa embeddings. The existence and accessibility of quality manifolds may vary with a different architecture or embedding model, a different embedding dimension, etc. 

\section{Future Work}

While our 1B model provides a solid baseline, scaling 3B or 7B parameter models would determine if our ML classifier impact becomes even stronger with more capacity. It would be particularly interesting to run cooldown experiments: instead of stopping at the stable phase, we could introduce a short, high-quality annealing phase (last 10-20\% of tokens) to see if we can ``recover'' performance on formal tasks while keeping the benefits of our broader multilingual filtering. There is also a big opportunity to test zero-shot transfer on truly low-resource languages like Swahili or Urdu, where native data is so scarce that cross-lingual ``subsidy'' from high-resource languages is the only viable path forward.
Another ablation idea would be to also apply a similar technique to the positive samples fed to the classifier. Thanks to our experiments, we argue that selecting better positives and combining that with the Q3 strategy could lead to an even bigger performance boost. Finally, the rigidity of the sample count threshold for the ML (Q3) strategy warrants further investigation. Relying on a static count (e.g., $200,000$ samples) is likely suboptimal. Future work should explore adaptive criteria based on score distribution analysis, variance, or density heuristics to dynamically determine eligibility for Q3 sampling. This would optimize the trade-off between refining the decision boundary and preserving valid high-quality tokens in medium-resource languages.

\end{document}

%% file: math_commands.tex
\usepackage{amsmath,amsfonts,bm}

\def\eqref#1{equation~\ref{#1}}

\def\1{\bm{1}}

\DeclareMathAlphabet{\mathsfit}{\encodingdefault}{\sfdefault}{m}{sl}
\SetMathAlphabet{\mathsfit}{bold}{\encodingdefault}{\sfdefault}{bx}{n}

